\documentclass[twocolumn, 11pt]{article}

\usepackage{arxiv}

\usepackage[utf8]{inputenc} %
\usepackage[T1]{fontenc}    %
\usepackage{hyperref}       %
\usepackage{url}            %
\usepackage{booktabs}       %
\usepackage{amsfonts}       %
\usepackage{nicefrac}       %
\usepackage{microtype}      %
\usepackage{lipsum}		%
\usepackage{graphicx}

\usepackage[numbers]{natbib}

\usepackage{doi}

\usepackage{xcolor}
\usepackage{multirow}

\usepackage{amssymb}  %
\usepackage{amsmath}  %
\usepackage{subcaption}  %
\usepackage{booktabs}   %
\usepackage{colortbl}  %

\usepackage{amssymb}
\usepackage{amsmath}
\usepackage{amsthm}

\usepackage{xcolor}
\usepackage{subcaption}
\usepackage{booktabs}
\usepackage{colortbl}

\usepackage{algorithm}
\usepackage{algorithmicx}
\usepackage{algpseudocode}

\usepackage{placeins}

\newtheorem{theorem}{Theorem}
\newtheorem{lemma}{Lemma}

\usepackage{etoc}

\newcommand{\AppendicesPart}{
  \refstepcounter{part}
  \addcontentsline{toc}{part}{Appendices}
  \part*{Appendices} %
}

\newcommand\blfootnote[1]{%
  \begingroup
  \renewcommand\thefootnote{}%
  \footnotetext{#1}%
  \addtocounter{footnote}{-1}%
  \endgroup
}

\title{Intrinsic Dimensionality as a Model-Free \\Measure of Class Imbalance}

\author{ 
    \textbf{Çağrı Eser}\thanks{Corresponding author.}\\
	Department of Computer Engineering\\
	METU, Ankara, Turkey \\
	\texttt{cagri.eser@ceng.metu.edu.tr} \\
	\And
	\textbf{Zeynep Sonat Baltacı} \\
	LIGM\\
    Ecole des Ponts, Marne-la-Vallée, France\\
    \texttt{sonat.baltaci@enpc.fr} \\
	\authorrowbreak
	\textbf{Emre Akbaş}\equalcontrib\\
	Department of Computer Engineering\\
	METU, Ankara, Turkey \\
	\texttt{emre@ceng.metu.edu.tr} \\
	\And
	\textbf{Sinan Kalkan}\equalcontribmark \\
	Department of Computer Engineering\\
	METU, Ankara, Turkey \\
	\texttt{skalkan@ceng.metu.edu.tr} \\
}

\date{}

\renewcommand{\headeright}{Preprint}
\renewcommand{\undertitle}{Preprint}
\renewcommand{\shorttitle}{ID as a Model-Free Measure of Class Imbalance}

\hypersetup{
pdftitle={Intrinsic Dimensionality as a Model-Free Measure of Class Imbalance},
pdfsubject={cs.LG, cs.CV},
pdfauthor={Cagri Eser},
pdfkeywords={intrinsic dimensionality, class imbalance, long-tailed learning},
}

\begin{document}

\maketitle
\blfootnote{\\To appear in Neurocomputing.}

\begin{abstract}
Imbalance in classification tasks is commonly quantified by the cardinalities of examples across classes. This, however, disregards the presence of redundant examples and inherent differences in the learning difficulties of classes. Alternatively, one can use complex measures such as training loss and uncertainty, which, however, depend on training a machine learning model. Our paper proposes using data Intrinsic Dimensionality (ID) as an easy-to-compute, model-free measure of imbalance that can be seamlessly incorporated into  various imbalance mitigation methods. Our results across five different datasets with a diverse range of imbalance ratios show that ID consistently outperforms cardinality-based re-weighting and re-sampling techniques used in the literature. Moreover, we show that combining ID with cardinality can further improve performance. Our code and models are available at \url{https://github.com/cagries/IDIM}.
\end{abstract}


\section{Introduction}
\label{sec:intro}
\begin{figure}[t!]
  \centering
 \includegraphics[width=.99\linewidth]
                   {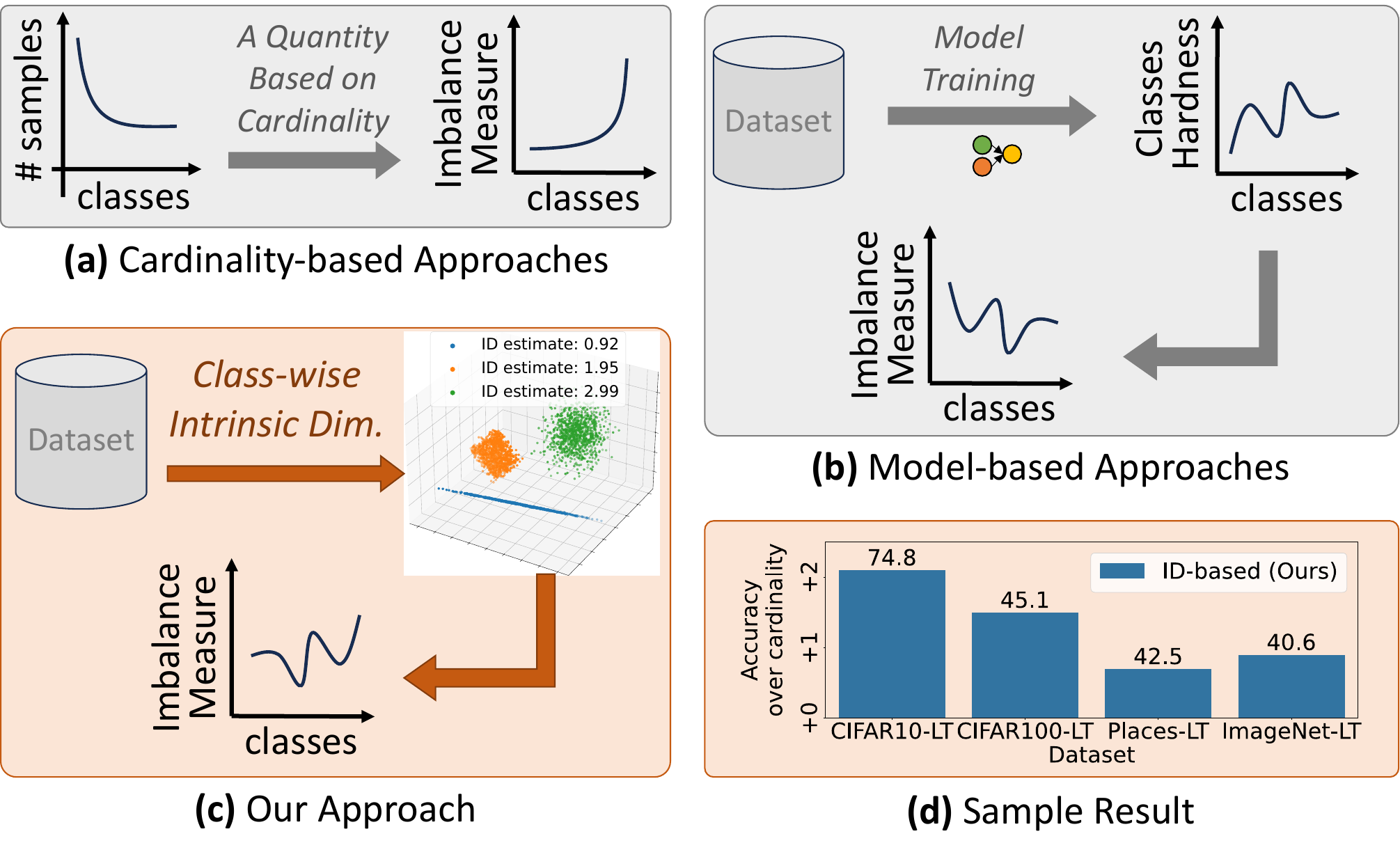}
  \caption{Existing approaches to long-tailed visual recognition rely on (a) the cardinalities of examples across classes, which is affected by redundant examples in the dataset, or (b) measures of class hardness, which require training an ML model. (c) Our approach of using data ID provides information about the manifold (and hence, complexity) of each class without being affected by redundant examples and without training an ML model. (d) We show that this simple, model-free, plug-and-play approach provides significant improvements with different approaches (a sample result shown here with a resampling approach).}
  \label{fig:id-teaser}
\end{figure}

Learning-based visual recognition is prone to differences among classes in training datasets. One aspect of such a difference is the cardinalities of examples across classes (called class cardinalities in short in the paper): Datasets generally exhibit a long-tailed distribution in terms of class cardinalities \citep{yang2022survey, zhang2023deep, menon2021longtail, cui2019class}, which has prompted research using class cardinality as a measure in various imbalance mitigation strategies \citep{drummond2003c4, kang2019decoupling, mahajan2018exploring} -- see also Fig. \ref{fig:id-teaser}a. 

Another aspect of imbalance that is more relevant to the learning dynamics involves the inherent discriminability or learning hardness of classes \citep{cui2019class, japkowicz2002class, cao2019learning, samuel2021distributional}. Although such hardness-based measures can be viewed as more established and promising, they require multiple training stages since a machine learning model needs to be trained first to quantify imbalance among classes (Fig. \ref{fig:id-teaser}b).

\begin{figure}[hbt!]
  \centering
       \includegraphics[width=1\linewidth]{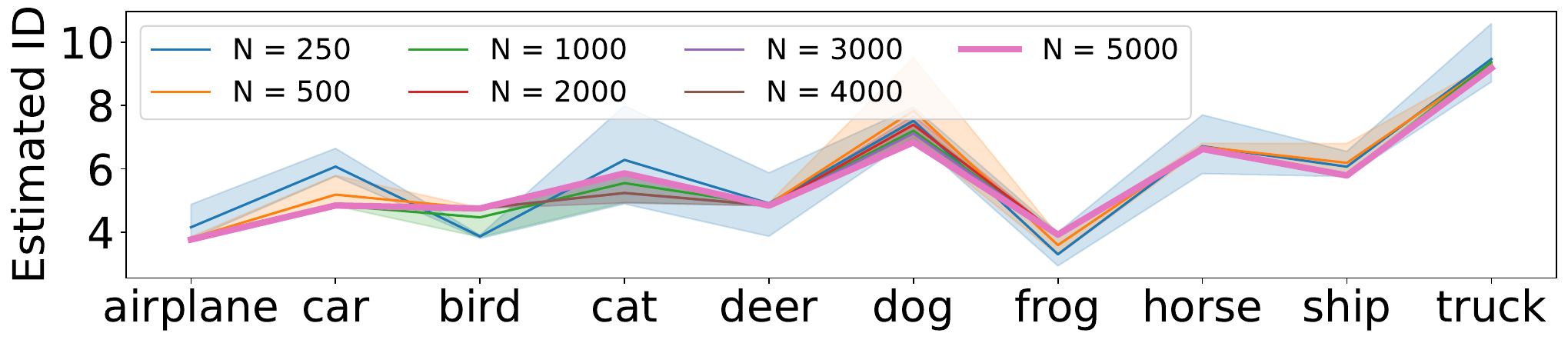}
       \caption{Estimated IDs for CIFAR-10 classes for different sample counts. ID can capture inherent differences among classes and is relatively robust against sample count, as exemplified here with the CIFAR-10 dataset, subsampled at different sample counts per class.}
       \label{fig:id-against-cardinality}
\end{figure}

In this paper, we introduce a model-free, cardinality-free measure of imbalance among classes based on data Intrinsic Dimensionality (ID) (Fig. \ref{fig:id-teaser}c). ID estimation \citep{fukunaga2013introduction, campadelli2015intrinsic, camastra2016intrinsic} is a well-established field of research that aims to estimate the implicit dimensions of lower-dimensional data manifolds embedded in a higher-dimensional space.

We first show that class-wise calculation of ID can point to inherent differences among classes (see, e.g., Fig. \ref{fig:id-against-cardinality}). For a given dataset, it is sufficient to compute the ID for each class once. Then, we integrate our ID-based measure into various imbalance mitigation strategies based on resampling, loss reweighting, and margin adjustment methods. We then compare ID-based imbalance mitigation against alternative approaches and evaluate ID-enhanced training combined with state-of-the-art methods. 

\noindent\textbf{Why Data ID?}
Manifold Learning Theory suggests that high-dimensional data often lie on a lower-dimensional manifold \citep{gorban2018correction, fefferman2016testing, pope2021intrinsic, ma2022delving}. Our results confirm that, in a multi-class dataset, the data manifold of each class has a specific ID \citep{ma2024unveiling}. Data ID quantifies this true dimensionality. Statistical learning theory \citep{vapnik1999overview} suggests that the number of \textit{effective} samples needed for a target  generalization error scales with (true) data dimensionality. Thus,  data ID for a class -- being independent of its sample size -- can act as a proxy for the amount of data required for modeling that class \citep{campadelli2015intrinsic}. 
Instead of relying only on the cardinality of classes, which completely disregards intrinsic properties of data, we argue that IDs of classes better capture the underlying structure of the data, leading to increased robustness in imbalance mitigation strategies, as we experimentally demonstrate.
Furthermore, our analysis on ID estimation (Section \ref{sect:analysis}) shows that data ID is robust against sample count, sample noise and extrinsic dimension,  qualities that we would expect from a measure of imbalance for long-tailed data distributions.

\noindent\textbf{Main Contributions}. (1) We introduce a novel measure of class imbalance based on data ID. We view this as a new perspective into the class imbalance literature as it complements existing measures based on cardinality and class hardness.
(2) We show that our ID-based measure can be easily integrated into different imbalance mitigation strategies.
(3) We demonstrate that our ID-based measure is able to quantify semantic imbalance among classes, which is missed by  cardinality-based methods.
(4) Across different datasets, we report significant improvements over using class cardinality as a measure. For example, incorporating data ID into progressively-balanced sampling increases accuracy by up to $2.1\%$ (Fig. \ref{fig:id-teaser}(d)) with virtually zero additional cost.

\section{Related Work}
\label{sec:relatedwork}

\subsection{Long-tailed Visual Recognition}

Long-tailed recognition approaches mitigate the effects of imbalance relying on sample counts or hardness under the following main categories \citep{yang2022survey, zhang2023deep, zhang2021bag}: Resampling methods \citep{kang2019decoupling, mahajan2018exploring}, loss-reweighting methods \citep{japkowicz2002class, cao2019learning, fernando2022dynamical, zhang2021distribution}, margin (logit) adjustment methods \citep{ren2020balanced, hong2021disentangling} and other methods \citep{cui2022reslt, alexandridis2024adaptive, zhang2022self}. For a detailed review, see Appendix \ref{sect:literature_long_tail} and, for formal definitions of the common approaches, see Section \ref{sec:using_ID}.

\subsection{Intrinsic Dimensionality}

Intrinsic dimensionality analysis focuses on quantifying the intrinsic qualities of data.
As a general definition, the intrinsic dimension is the minimum number of parameters required to represent a data distribution with minimal loss of generality \citep{fukunaga2013introduction}.
ID is closely related to the manifold hypothesis \citep{gorban2018correction}, which posits that high dimensional natural image datasets  lie on a lower dimensional manifold, enabling learning.
In this context, ID refers to the number of dimensions of this lower-dimensional manifold. The ID of a dataset can be estimated globally (i.e., for the entire dataset), or locally (around a local neighborhood), depending on the scope and the estimation method.
ID estimation methods in the literature can be studied in two general categories: Projective methods \citep{bishop1998bayesian,albergante2019estimating,roweis2000nonlinear,tenenbaum2000global}  and geometric (topological) methods \citep{levina2004maximum, amsaleg2019intrinsic, facco2017estimating}. For the sake of space, we provide a further review of these methods in Appendix \ref{sect:literature_ID}.

A recent work by \citet{ma2024unveiling} focuses on the intrinsic dimension of perceptual manifolds of balanced datasets in the context of model fairness and introduces ID regularization as an end-to-end trainable loss, and similarly the work by \citet{ansuini2019intrinsic} also studies the ID of the representation of data in the intermediate layers of a neural network. In comparison, we introduce ID as an imbalance measure and show that it can be used for training as a model-free mitigation method in the context of imbalanced datasets.

\subsection{Comparative Summary}

We observe that the class imbalance literature has mainly used class cardinality or class hardness (obtained through some measure of learning difficulty, e.g., prediction probability, class loss). To our knowledge, we are the first to introduce data ID as a model-free measure of class imbalance and use it in various imbalance mitigation strategies.

\section{FisherS: Background on ID Estimation}
\label{sec:id}

We first introduce the (data) Intrinsic Dimension (ID) estimation method we have used, the method of \citet{albergante2019estimating}, which we call FisherS, for estimating the IDs of individual classes. As we will analyze through experimentation, our approach and conclusions are not dependent on the specific ID estimation method.

In essence, FisherS utilizes the concept of \textit{blessing of dimensionality} where points from samples with high dimensionality can be separated from the rest of the sample set by linear inequalities with high probability, regardless of the sample count \citep{gorban2018correction}.
With some preprocessing, these linear inequalities can be estimated by Fisher's linear discriminant in an efficient manner. 

For a dataset $X$ with points $\mathbf{x} \in \mathbb{R}^{n}$, the FisherS ID is estimated with the following steps:

\noindent\textbf{Preprocessing}. The data is initially preprocessed by centering and projecting onto the linear subspace spanned by its first major principal components found by PCA \citep{fukunaga1971algorithm} to reduce collinearity in the data. The major components are defined as components whose eigenvalues are higher than $\frac{\lambda_{max}}{C}$, with $\lambda_{max}$ being the largest eigenvalue of the sample covariance matrix, and $C$ a constant (a good value being $C = 10$ to reduce collinearity \citep{gorban2018correction, 87e6ea2d996a4ec09e00c4a4c4e9b9a0}).
Afterwards, whitening (i.e., linearly transforming data points such that the covariance matrix is an identity matrix) is applied, and the obtained vectors are normalized to unit length so that each vector is a projection into a unit sphere.

\noindent\textbf{Fisher Separability}. Given a dataset preprocessed as described above, a point $\mathbf{x} \in X$ is said to be Fisher-separable from a collection of points $Y$ with parameter $\alpha \in [0, 1)$ if:
\begin{equation}
\label{eqn:albergante0}
    \langle \textbf{x}, \textbf{y} \rangle \leq \alpha \langle \textbf{x}, \textbf{x}\rangle,
\end{equation}
for all $\mathbf{y} \in Y$, with $\langle\mathbf{x}, \mathbf{y}\rangle = \sum^{n}_{k=1}x_{k}y_{k}$ being the inner product of $\mathbf{x}, \mathbf{y} \in \mathbb{R}^{n}$ \citep{gorban2018correction}.
Similarly, the dataset $X$ is Fisher-separable with parameter $\alpha$ if~(Eq. \ref{eqn:albergante0}) is true for all $\mathbf{x} \in X$.

\noindent\textbf{FisherS ID Estimation}. Let $p_{\alpha}(\mathbf{y})$ be the probability that a point $\mathbf{y}$ is not separable from all other points, and $\bar{p_{\alpha}}(\mathbf{y})$ be the mean value of the distribution of $p_{\alpha}(\mathbf{y})$ over all data points.  With the data projected onto a unit sphere, \citet{albergante2019estimating} show that $p_{\alpha}(\mathbf{y})$ can be estimated on the rotationally invariant equidistribution on the unit sphere $\mathbb{S}^{n-1} \subset \mathbb{R}^{n}$ by:
\begin{equation}
\label{eqn:albergante1}
    p_{\alpha} \approx \frac{1}{\alpha \sqrt{2\pi n}}(1 - \alpha^{2})^\frac{n-1}{2},
\end{equation}
where the distribution of $p_{\alpha}$ is a delta function centered in $\bar{p_{\alpha}}$. 
In this case, the ID $n_{\alpha}$ of the data for parametrized $n_{\alpha}$ can be obtained by solving for $n$ in~(\ref{eqn:albergante1}) as:
\begin{equation}
\label{eqn:albergante2}
    n_{\alpha} = \frac{1}{-\ln(1 - \alpha^{2})} W\left(\frac{-\ln(1 - \alpha^{2})}{2\pi \bar{p}_{\alpha}^{2}\alpha^{2}(1 - \alpha^{2})}\right),
\end{equation}
where $W(\cdot)$ is the Lambert function. In FisherS, this process is repeated for many values of $\alpha$, and the $\alpha$ parameters with valid $n_{\alpha}$ dimensions are kept in $\boldsymbol{\alpha}_{s}$. The $n_{\alpha}$ corresponding to $\alpha$ that is closest to $0.9 \times \max\{\boldsymbol{\alpha}_{s}\}$ is determined to be the intrinsic dimension, $d$, of the dataset $X$.

\noindent\textbf{Our Application}. In our study, we estimate class-wise data ID for each class $c$, denoted as $d_c$, by applying the aforementioned FisherS ID estimation to the (training) samples of class $c$. Note that we do this \textit{once per dataset, without requiring any form of deep model training beforehand}. 
Our method  relies on the relative differences of ID estimates between classes. 
As a result, for our mitigation experiments, we normalize $d_c$ to obtain $\hat{d}_c$ so that $\sum_c \hat{d}_c = 1$:
\begin{equation}
    \hat{d}_c = \frac{\textrm{FisherS}(\{\mathbf{x}_i\ |\ y_i = c \})}{\sum_{c'} \textrm{FisherS}(\{\mathbf{x}_i\ |\ y_i = c' \})}.
    \label{eq:classwise_id}
\end{equation}

\begin{figure}[hbt!]
    \begin{subfigure}[b]{0.9\linewidth}
     \centering
       \includegraphics[width=\linewidth]{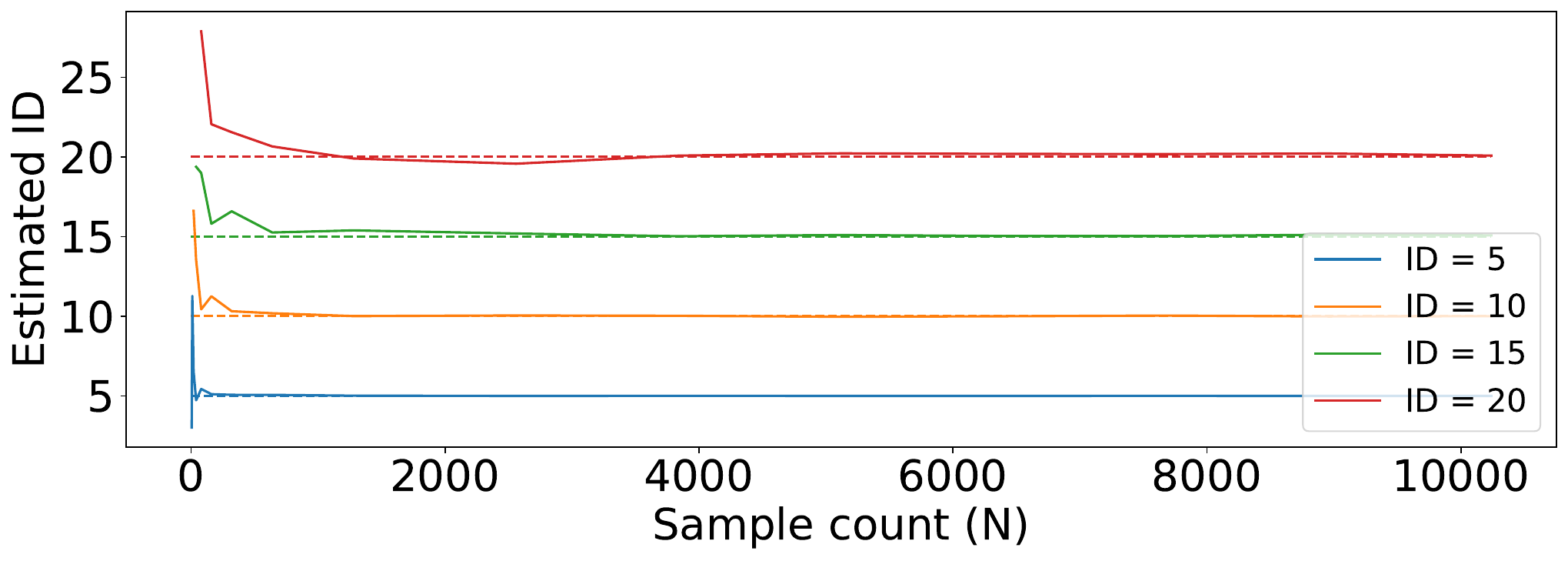}
       \caption{ID vs. sample count [synthetic data].}
       \label{fig:id-against-sample-size}
    \end{subfigure}
    \begin{subfigure}[b]{0.9\linewidth}
     \centering
       \includegraphics[width=\linewidth]{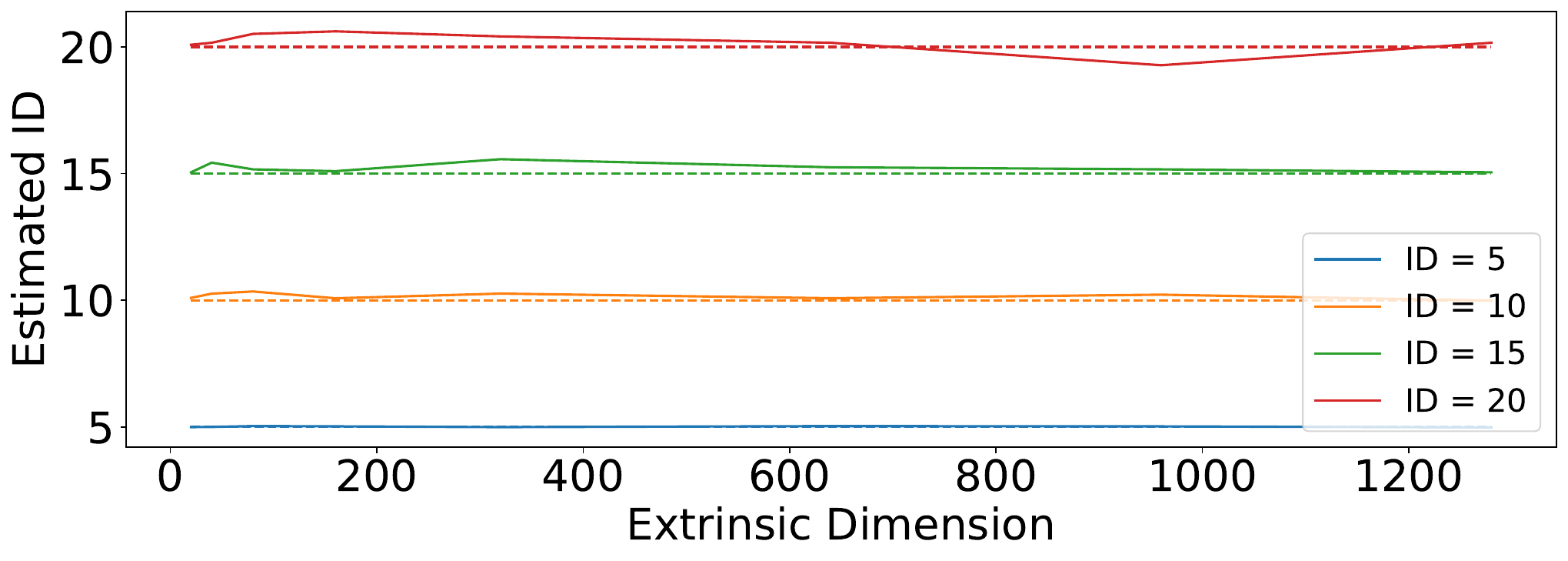}
        \caption{ID vs. extrinsic dimension [synthetic data].}
    \label{fig:id-against-extrinsic-dimension}
    \end{subfigure}
    \caption{Analysis on FisherS estimated IDs. The analysis with the synthetic data in (a) and a real dataset in Fig. \ref{fig:id-against-cardinality} suggest that ID estimation is robust against sample count. Moreover, the analysis in (b) shows that it is not affected by extrinsic dimensionality either. Dashed lines: true ID values.
    }
    
\end{figure}

\section{ID as a Measure of Class Imbalance}
\label{sect:analysis}

\noindent\textbf{Why an ID-based imbalance measure?} Class imbalance analysis and mitigation methods require quantifying the amount of imbalance across classes. As discussed in Introduction, data ID by definition provides a measure of the true dimensionality (complexity) of the data manifold. Statistical learning theory \citep{vapnik1999overview} suggests that the number of \textit{effective} samples needed for a target  generalization error scales with (true) data dimensionality (complexity). Thus,  data ID for a class -- being independent of its sample count -- captures the complexity (true dimensionality) of that class and as suggested by statistical learning theory, provides a proxy for the necessary number of samples required for modeling (learning) that class \citep{campadelli2015intrinsic}. This motivates and justifies our use of data ID with resampling or reweighting methods. Similarly, a higher ID implies a more complex manifold, potentially requiring a decision boundary with wider margins, which justifies using data ID in logit adjustment. These arguments align with prior work on sample requirements  in imbalanced settings, such as  the  manifold-based derivation of the ``effective number of samples'' \citep{cui2019class} and the feature covariance based measure \citep{ma2022delving}. 

\noindent\textbf{ID estimation is robust against sample count.}
Following similar analysis \citep{hein2005intrinsic}, we generate synthetic data by sampling from a multivariate normal distribution. For ID = $d$, we sample from $\mathcal{N}(\boldsymbol{\mu}, \mathbf{\Sigma})$, where $\boldsymbol{\mu}$ is a $d$-dimensional zero vector and $\mathbf{\Sigma}$ is a $d \times d$ identity covariance matrix. Our results in Fig.~\ref{fig:id-against-sample-size} indicate that FisherS ID is robust against changing sample count,
which is very important when being used with long-tailed data distributions.
Furthermore, in Fig. \ref{fig:id-against-cardinality} we calculated the ID estimates of classes on randomly sampled subsets of the real-world CIFAR-10 dataset and observed the resulting ID estimates are relatively robust under a variety of example counts, which reflects this quality on real-world data.

\noindent\textbf{ID estimation is robust against extrinsic dimensionality.}
For this, we follow a similar approach to \citet{hein2005intrinsic} and generate synthetic data from multivariate normal distributions with fixed intrinsic dimensions, and progressively increase the extrinsic dimension of the data. For ID = $d$, we sample from $\mathcal{N}(\boldsymbol{\mu}, \mathbf{\Sigma})$, where $\boldsymbol{\mu}$ is a $D$-dimensional zero vector and $\mathbf{\Sigma}$ is a $D \times D$ covariance matrix with
\begin{equation}
  \mathbf{\Sigma} = \left[\begin{array}{ c  c }
    I_{d \times d} & 0 \\
    0 & 0
  \end{array}\right]_{D \times D}.
\end{equation}
After sampling, we rotate the generated samples around the origin \citep{vonLindheim2018} by applying pairwise rotations for each consecutive pair of dimensions, eliminating constant valued dimensions. Fig.~\ref{fig:id-against-extrinsic-dimension} shows that ID estimation is robust to changes in extrinsic dimension.

\begin{figure}[ht!]
     \centering
   \includegraphics[width=\linewidth]{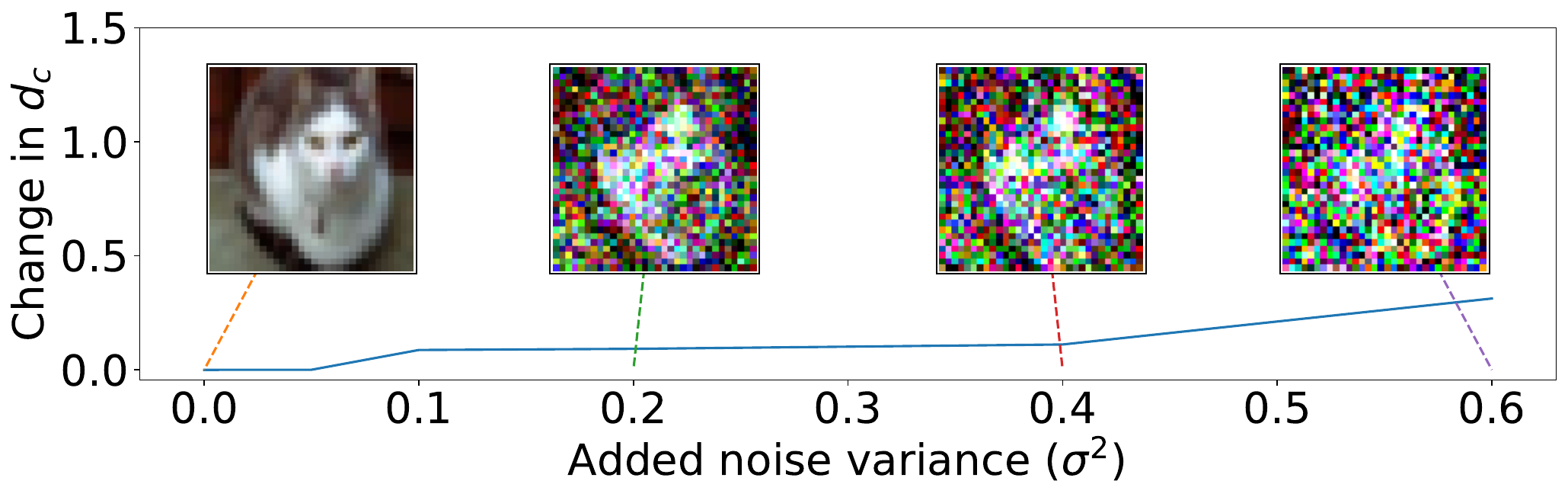}
   \caption{ID estimates of CIFAR-10 classes with added Gaussian noise of varying scales ($\sigma\in[0, 1]$), showing robustness of ID to even drastic amounts of noise.}
   \label{fig:id-with-increasing-noise}
\end{figure}

\noindent\textbf{ID estimation is robust against data noise.}
To investigate the effect of noise for estimating ID in real-world high-dimensional data, we added Gaussian noise with varying scales to CIFAR-10 images (channel intensities normalized between $[0, 1]$) and calculated their class-wise ID. Fig. \ref{fig:id-with-increasing-noise} indicates that ID estimation is robust against even extreme levels of noise where the images are hardly identifiable.

\noindent\textbf{ID estimation is robust against correlation.} We also show that ID estimation is relatively robust against dimension correlation.
See Appendix \ref{sect:robustness_analysis} for this analysis.

\subsection{Theoretical Motivation}

To theoretically motivate our use of class-wise ID as a suitable measure of imbalance, we follow the class-wise manifold analysis work by Narayanan \& Niyogi \cite{narayanan2009sample}. Narayanan \& Niyogi proved that the number of balls of radius $\epsilon$ required to cover the space of smooth boundaries on a $d$-dimensional manifold is related to the manifold's volume and intrinsic dimension and to sample complexity.

Let the total support of the data $X$ be the union of $K$ disjoint compact Riemannian sub-manifolds, where each manifold $\mathcal{M}_k$ corresponds to a class indexed by $k$:
\begin{equation}
    X = \bigcup_{k=1}^{K} \mathcal{M}_k.
\end{equation}
Each class manifold $\mathcal{M}_k$ has its own distinct intrinsic dimension $d_k$.  Moreover, let $R_k(f) = \mathbb{E}_{x \sim \mathcal{M}_k} [\mathbb{I}(f(x) \neq k)]$ denote the generalization error specifically on class $k$.  %

\begin{theorem}[Class-wise Intrinsic Dimension and Sample Complexity] 
The number of samples $n_k$ for class $k$ required to bound the generalization error $R_k(f) \le \epsilon$ with high probability is:
\begin{equation}
    n_k \gtrsim \Omega\left( \left(\frac{1}{\epsilon}\right)^{d_k} \right).
\end{equation}
\end{theorem}

\begin{proof}
See Appendix \ref{sect:proof} for a sketch of the proof.
\end{proof}

This theorem suggests that a class with low ID ($d_k$ small) is easy to learn even with few samples. On the other hand, a class with high ID ($d_k$ large) requires exponentially more samples. If we calculated a single ``global'' ID for the whole dataset, we would average out these differences, potentially masking the difficulty of a complex minority class.

Motivated by the theoretical link between class-wise ID and sample complexity, we propose using class-wise ID ($d_k$) as a measure of class imbalance, which we can use in place of class cardinality in designing sampling-based, loss-reweighting-based, or logit-adjustment-based imbalance mitigation strategies. However, if a class is mostly noise, lacking sufficient structure, its ID estimation will be high, which can negatively affect the imbalance mitigation strategy. This, on the other hand, is assumed to be rare in practice.

\section{Using ID to Mitigate Class Imbalance}
\label{sec:using_ID}

In this section, we describe how we use per-class ID estimates, $\hat{d}_c$ (Eq. \ref{eq:classwise_id}), in existing imbalance mitigation methods. The underlying principle is that the intrinsic dimension of more complex classes is expected to be higher than that of simpler classes, which should be easier to learn. 

Specifically, for resampling-based experiments, we use $\hat{d}_c$ for a dataset to specify the sampling probabilities for class $c$, and then uniformly sample from the samples of class $c$. 
The higher the estimated ID is for a specific class category, the more ``difficulty'' it poses for a trained model to learn. Therefore we propose that this class should be sampled more frequently for the trained model to capture the intrinsic qualities of the class more effectively. 
Similarly, when plugging ID-based learning into reweighting methods, the intuitive idea is to have lower loss weighting for ``simpler'' classes (i.e.,  classes with lower ID estimates). 
We do not make use of any sample-size assumptions here and completely rely on the intrinsic qualities captured by ID in this approach.

\noindent\textbf{Resampling methods.}
In standard sampling, each sample in the training dataset has an equal probability to be sampled, i.e., sampling from a class is directly correlated with the class cardinality of that class \citep{yang2022survey, zhang2023deep, chawla2002smote}: 
\begin{equation}
 p_{\mathbf{x}} = p_{c} \times \frac{1}{N_{c}},
\end{equation}
where $p_{\mathbf{x}}$ is the probability of sampling a training sample $\mathbf{x}$ of class $c$ and $N_{c}$ the number of samples in class $c$. Resampling methods define sampling probability for class $c$, $p_c$ as:
\begin{equation}
 p_{c} = \frac{N_{c}}{\sum_{c'} N_{c'}}.
\end{equation}

\underline{Our Modification}. In our method, we initially sample the class category $c$ by using the estimated ID of class $c$, and then uniformly sample from the class samples:
\begin{equation}
    p_{c} = \frac{\hat{d}_{c}}{\sum_{c'}\hat{d}_{c'}}.
\end{equation}

\noindent \textbf{Loss Reweighting methods.}
Class-based loss reweighting methods scale the loss individually for each class~\citep{japkowicz2002class, cui2019class}:
\begin{equation}
    \mathcal{L} = \sum_{c} w_{c}\mathcal{L}_{c},
\end{equation}
where $\mathcal{L}_{c}$ is the loss for the samples from class $c$ and $w_{c}$ is the loss weight assigned to samples from class $c$. $w_{c}$ is assigned some values based on the cardinality of samples in class $c$; e.g.,  $w_{c} = \frac{n_{\textrm{min}}}{n_{c}}$ \citep{japkowicz2002class}, with $n_{\textrm{min}}$ being the cardinality of the least populated class and $n_{c}$ being the cardinality of class $c$.

\underline{Our Modification}. In our method, we substitute $w_{c}$ by $\hat{d}_{c} \times |\mathcal{C}|$, where $\mathcal{C}$ is the set of classes.

\noindent\textbf{Margin-based methods.}
Since margin-based methods use different strategies, we integrate $\hat{d}_c$ on a case-by-case basis. 

In the case of \textbf{LDAM} \citep{cao2019learning}, we use the LDAM + deferred re-weighting framework of \citep{zhang2021bag}.
Specifically, LDAM loss has the following structure:
\begin{align}
    \mathcal{L}_\mathrm{LDAM}(\mathbf{x},y) &= -\log\frac{\exp({z_{y} - \Delta_{y}})}{\exp({z_{y} - \Delta_{y} + \sum_{c \neq y} \exp({z_{c}})})}, \nonumber \\
    \Delta_{c} &= \frac{C}{N_{c}^{\frac{1}{4}}}, 
\end{align}
where $z_{c}$ denotes the output of the model for class $c$, $\Delta_{c}$ is the margin for class $c$, and $C$ is a constant.

\underline{Our Modification}. Similar to \citet{baltaci2023class}, we use the ID estimate for each class by setting the margin as:
\begin{equation}
    \Delta_{c} = 0.5 \times \frac{\hat{d_{c}}}{\max_{c}\hat{d_{c}}}.
\end{equation}

\textbf{DRO-LT} \citep{samuel2021distributional} adjusts the margin with a more robust measure:
\begin{align}
\mathcal{L}_\textrm{DRO}(\mathbf{x}, y) &= - w(y) \log \frac{\exp({-d(\hat{\mu}_{y}, \mathbf{x}) - \Delta_{y}})}{\sum_{\mathbf{x}'}\exp({-d(\hat{\mu}_{y}, \mathbf{x}') - \Delta'_{y}})}, \nonumber \\
\Delta_{y} &= 2 \varepsilon_{y},
\end{align}
where $d(\hat{\mu}_{c}, z)$ is the distance in the feature space between the sample $\mathbf{x}$ and the estimated centroid of its class  $\hat{\mu}_{y}$, and $\Delta_{y}$ is the class margin for class $y$. 

\underline{Our Modification}. Similar to the case of LDAM loss, we utilize class ID estimates by assigning the margin as: 
\begin{equation}
    \Delta_{c} = \frac{\hat{d_{c}}}{\sum_{c'}\hat{d_{c'}}} \times C. \\
\end{equation}

In the case of learnable margins in DRO-LT Loss, we initialize the margin values $\varepsilon_{c}$ with the normalized ID estimates $\hat{d}_{c}$ and allow these epsilon values to update during training as done in the original paper.

The \textbf{Post-hoc logit adjustment} method proposed by \citet{menon2021longtail}, predicts a label according to:
\begin{equation}
    \operatorname*{argmax}_{y}\frac{\exp(w_{y}^{\top}\Phi(x))}{\pi_{y}^{\tau}} = \operatorname*{argmax}_{y}f_{y}(x) - \tau \log \pi_{y},
\end{equation}
where $f_{y}(x) = w_{y}^{\top}\Phi(x)$ represents the logits of a neural network, $\tau$ a tuning  parameter and $\pi \in \Delta_{y}$ are estimates of the class priors $p(y)$. 

\underline{Our Modification}. 
Instead of using the empirical class frequencies for $\Delta_{y}$, we adopt $\hat{d}_y$ as follows: 
\begin{equation}
  \Delta_{y} = \frac{1/\hat{d}_{y}}{\sum_{c'} 1/\hat{d}_{c'}}. 
\end{equation}

\section{Experiments and Results}
\label{sec:experiments}

\textbf{Datasets}. We evaluate our method through the CIFAR-10-LT and CIFAR-100-LT, Places-LT and ImageNet-LT datasets that are commonly used in the long-tailed visual recognition literature, as well as the semantically imbalanced SVCI-20 dataset.
 For details about the datasets, see Appendix \ref{sect:dataset_details}.

\noindent \textbf{Evaluation Measure.}
As in prior work, we use the top-1 classification accuracy: $N^{v}_{t}/N^{v}$, 
where $N^{v}$ is the number of examples in the validation set, and $N^{v}_{t}$ is the number of correctly-predicted examples.

\noindent\textbf{Training \& Implementation Details} are provided in Appendix \ref{sect:imp_and_training_details}.

\begin{table}[hbt!]
    \centering
    \small

    \begin{tabular}{l | c c | c c }
		\toprule 
        \multicolumn{1}{c}{Dataset:}       & \multicolumn{2}{c}{CIFAR-10-LT} & \multicolumn{2}{c}{CIFAR-100-LT} \\
		\midrule 
        \hfill Imbalance Ratio ($\rho$):                              & 100                              & 50                             & 100             &  50            \\
        \midrule
        \rowcolor{gray!30!}
        \multicolumn{5}{c}{Reweighting-based methods} \\
		\midrule
        CE (baseline)\hfill\                                & 71.6 & 76.3                                                      & 38.9 & 42.9                    \\
	CB + Focal\hfill\                                             & 73.9 & 79.5                                                      & 38.4 & 43.1                    \\
     CB + Focal [2S]\hfill\                                & 74.4 & 78.8                                                      & 38.0 & 44.5                    \\
	CSCE \citep{japkowicz2002class} [2S]\hfill\            & 74.7 & 79.2                                                      & 41.1 & 45.4                    \\
        \midrule
	CB + Focal + ID [\textbf{Ours}]\hfill\                & 74.8 & 78.7                                                      & 40.8 & 43.0                    \\
	ID [\textbf{Ours}] [2S]\hfill\                         & \textbf{76.4} & \textbf{80.8}                                    & \textbf{41.7} & \textbf{46.2}  \\
		\midrule
        \rowcolor{gray!30!}
        \multicolumn{5}{c}{Resampling-based methods}  \\
		\midrule
        CE (baseline)\hfill\                                 & 71.6 & 76.3                                                      & 38.9 & 42.9                    \\
	PB \citep{kang2019decoupling}\hfill\                           & 72.7 & 77.8                                                      & 39.1 & 43.6                    \\
	PB \citep{kang2019decoupling} [2S]\hfill\              & 66.5 & 75.4                                                      & 38.7 & 43.1                    \\
	CB sampling \citep{mahajan2018exploring} \hfill\                            & 72.3 & 77.0                                                     & 33.4 & 40.4                    \\
	CB sampling \citep{mahajan2018exploring} [2S]\hfill\              & 71.0 & 78.7                                                      & 40.2 & 44.3                    \\
        \midrule
        ID [\textbf{Ours}]\hfill\                            & 71.1 & 78.5                                                      & 35.2 & 39.4                    \\
	PB-ID [\textbf{Ours}]\hfill\                                & 74.4 & \textbf{78.9}                                             & 39.7 & \textbf{45.1}           \\
	PB-ID [\textbf{Ours}] {[2S]}\hfill\                   & \textbf{74.8} & 78.8                                             & \textbf{40.8} & 44.8           \\
        \midrule
        BoT \citep{zhang2021bag} [2S]\hfill\  & 80.0 & 83.6                                                      & 47.8 & 51.7                    \\
	BoT + ID [\textbf{Ours}] [2S]\hfill\         & \textbf{81.8} & \textbf{85.1}                                    & \textbf{48.6} & \textbf{53.1}  \\
		\bottomrule
	\end{tabular}
 	\caption{\textbf{Experiment 1:} Top-1 accuracy results for plug-and-play reweighting and resampling methods for ResNet-32
	on long-tailed CIFAR datasets. [2S]: Two-stage methods.}
	\label{tab:cifar-plug-and-play} 
\end{table}

\subsection{Experiment 1: Model-Free Approaches}
We first evaluate the performance of ID-based imbalance mitigation against alternative plug-and-play model-free class imbalance mitigation methods on several datasets.

\textbf{Re-sampling and Re-weighting}. 
The results on the CIFAR-LT datasets in Table~\ref{tab:cifar-plug-and-play} show that both ID-based resampling and ID-based reweighting methods outperform alternative methods.
Furthermore, we observe that our method keeps its effectiveness under increasing data imbalance,
in our detailed analysis of one-stage methods in Appendix \ref{sect:additional_results}.

\textbf{Large-scale Datasets}. Similarly, our experiments with the Places-LT and ImageNet-LT datasets indicate that ID-based mitigation is also effective with larger-scale datasets (see Table~\ref{table:id-comparison-places-and-imagenet-results}). 
In particular, ID-based resampling scores at least $0.8\%$ higher than the top re-sampling methods on both datasets, while increasing accuracy by at least $1.1\%$ compared to the top re-weighting methods.

\textbf{Margin Adjustment}. Aside from reweighting and resampling, we also experiment with integrating ID into margin-based methods on a case-by-case basis in Table~\ref{table:cifar-margin} (additional results on CIFAR-100 can be found in Appendix \ref{sect:additional_results}). In this case, ID-based mitigation performs better in some cases while providing second-best results in others. 

\noindent\textbf{Summary}. Overall, our results indicate that using data ID ($\hat{d}_c$) provides better results than cardinality-based approaches and performs on par with learnable margin-based approaches. Being model-free, our measure stands out as an easy-to-use alternative measure of class imbalance.

\begin{table}[hbt!]
	\centering
    \small
	\begin{tabular}{l | c | c }
		\toprule \multicolumn{1}{c|}{Dataset:}                                      & Places-LT                         & ImageNet-LT    \\
		\midrule \multicolumn{1}{c|}{Backbone:}                                     & ResNet-152                        & ResNet-10      \\
		\midrule
        \rowcolor{gray!30!}
        \multicolumn{3}{c}{Reweighting-based methods} \\
        \midrule
          CE (baseline)\hfill\                                 & 32.68  & 34.01          \\
	   Focal Loss \citep{lin2017focal} \hfill\                         & 31.10          & 32.64          \\
	   CB \citep{cui2019class} \hfill\                                 & 39.44          & 33.50          \\
	   CB \citep{cui2019class} {[2S]}\hfill\                    & 40.44          & 39.27           \\
         CSCE \citep{japkowicz2002class} {[2S]}\hfill\  & 40.70 & \underline{39.43} \\

	CB + Focal\hfill\                                             & 37.36         & 31.04          \\
		CB + Focal {[2S]}\hfill\                                & \textbf{41.02} & 35.88           \\
		\midrule
         ID [\textbf{Ours}]\hfill\                            & 35.35          & 37.11           \\
		ID [\textbf{Ours}] {[2S]}\hfill\                        & \underline{40.91}          & \textbf{39.87}   \\
	CB + Focal + ID [\textbf{Ours}]\hfill\                        & 39.13          & 37.26           \\
	    CB + Focal + ID [\textbf{Ours}] {[2S]}\hfill\           & 39.09          & 38.82           \\
		\midrule
        \rowcolor{gray!30!}
        \multicolumn{3}{c}{Resampling-based methods}  \\
		\midrule 
    CE (baseline)\hfill\                                 & 32.68          & 34.01          \\
	PB \citep{kang2019decoupling}\hfill\                           & 40.45          & 38.48          \\
  	PB \citep{kang2019decoupling} {[2S]}\hfill\              & \underline{41.87}          & \underline{39.74}          \\
	CB \citep{mahajan2018exploring}\hfill\                           & 40.01          & 34.16          \\
	CB \citep{mahajan2018exploring} {[2S]}\hfill\              & 40.79          & 39.27          \\
		\midrule
       ID [\textbf{Ours}]\hfill\                            & 38.12          & 33.27           \\
	ID [\textbf{Ours}] {[2S]}\hfill\                        & 41.65          & 39.55           \\
	PB-ID [\textbf{Ours}] \hfill\                               & 38.52          & 38.78           \\
	PB-ID [\textbf{Ours}] {[2S]}\hfill\                   & \textbf{42.62 \scriptsize{$\pm$ 0.15}} & \textbf{40.55\scriptsize{$\pm$ 0.09}}\\
		\midrule
       BoT \citep{zhang2021bag} \hfill\            & 43.07          & \textbf{43.16} \\
	BoT + ID [\textbf{Ours}]\hfill\                     & \textbf{43.39 \scriptsize{$\pm$ 0.09}} & 42.92 \scriptsize{$\pm$0.11}         \\
		\bottomrule
	\end{tabular}
	\caption{\textbf{Experiment 1:} Top-1 accuracy results for plug-and-play reweighting and resampling methods on
	Places-LT and ImageNet-LT. Two-stage methods marked with [2S].}
	\label{table:id-comparison-places-and-imagenet-results}
\end{table}

\begin{table}[hbt!]
 \centering
 \small
 
 \setlength{\tabcolsep}{1mm}
\begin{tabular}{l | c c c  }
		\toprule 
        \multicolumn{1}{c|}{Dataset:}    & \multicolumn{3}{c}{CIFAR-10-LT} \\
        \midrule
        \multicolumn{1}{c|}{Imbalance Ratio ($\rho$):}                                    & 10                              & 50                              & 100                     \\
		\midrule\midrule CE (baseline)\hfill\                                                   & 87.00                           & 76.29                           & 71.58                   \\
		LDAM + DRW \citep{cao2019learning}\hfill\                            & 87.76                           & 81.64                           & 77.19          \\
		Log. Adj. \citep{menon2021longtail}\hfill\                       & -                               & -                               & 77.30             \\
		DRO-LT (learned $\epsilon$) \citep{samuel2021distributional} \hfill\  & \underline{91.02}                           & \underline{85.88}                           & \textbf{82.39}  \\
		\midrule LDAM + DRW + ID [\textbf{Ours}]\hfill\                     & 87.81                           & 81.12                           & 76.39              \\
		Log. Adj. + ID ($\tau = \tau$*) [\textbf{Ours}]\hfill\       & -                               & -                               & 74.10 \\
		DRO-LT + ID ($\varepsilon =$ N. ID) [\textbf{Ours}]\hfill\  & 90.56                           & 85.39                           & 81.73          \\
		DRO-LT + ID ($\varepsilon =$ L. + ID) [\textbf{Ours}]\hfill\   & \textbf{91.10}                  & \textbf{85.91}                  & \underline{82.34}            \\
		\bottomrule
	\end{tabular}
 	\caption{\textbf{Experiment 1:} Top-1 accuracy results for simple plug-and-play margin-based methods on long-tailed CIFAR-10. N.: normalized. L: Learned epsilon. Log. Adj.: Logit Adjustment. See Appendix \ref{sect:additional_results} for the results on CIFAR-100-LT.}
	\label{table:cifar-margin}
\end{table}
\subsection{Experiment 2: SOTA Comparison}
In these experiments, we integrate ID into state-of-the-art (SOTA) methods and evaluate its performance. Following the success of ID-based resampling in Experiment 1, we integrate ID-based resampling to the methods in these experiments with virtually no additional cost.

The results in Table \ref{tab:cifarlt-sota} on CIFAR-LT suggest that, with Bag of Tricks \citep{zhang2021bag} + ID, we boost the performance of the original model in all categories, while our GLMC \citep{du2023global} + ID method surpasses the state-of-the-art for CIFAR-LT classification in two of the four long-tail categories.
In the case of Places-LT (Table \ref{table:placeslt-sota}), ID-based resampling combined with Bag of Tricks \citep{zhang2021bag} again achieves the best accuracy with a low variance. 
On ImageNet-LT, we compare methods using ResNet-50 and ResNeXt-50 backbones following prior work and show the results in Table~\ref{tab:imagenetlt-sota}. We see that ID-based resampling boosts the performance of both Bag of Tricks \citep{zhang2021bag} and BCL \citep{zhu2022balanced} with ResNet-50 backbones, with the latter achieving state-of-the-art results for a single model under 100 epochs.
In the case of ResNeXt-50 backbones, BCL \citep{zhu2022balanced} + ID again improves the accuracy of the original BCL method, and obtains the best accuracy under 100 epochs.

\noindent\textbf{Summary}. Our results indicate that data ID can complement and improve existing imbalance mitigation strategies, achieving SOTA results on several datasets.

\begin{table*}[hbt!]
   \small
   \centering
   \begin{tabular}{l  r | c	c | c c }
		\toprule
        \multicolumn{2}{c}{Dataset:}  & \multicolumn{2}{c}{CIFAR-10-LT} & \multicolumn{2}{c}{CIFAR-100-LT} \\
		\midrule \multicolumn{2}{c|}{Imbalance Ratio ($\rho$): }    & 100      & 50        & 100     &  50     \\
        \midrule\midrule
        CB + Focal & - & 74.6 & 78.8 & 38.0 & 44.5 \\
        BBN \citep{zhou2020bbn} & CVPR'20 & 79.8 & 82.2 & 44.1 & 49.2 \\
        BALMS \citep{ren2020balanced} & NeurIPS'20 & 84.9 & - & 50.8 & 54.2 \\
		BoT \citep{zhang2021bag} & AAAI'21 & 80.0 & 83.6 & 47.8 & 51.6 \\
        MiSLAS \citep{zhong2021improving} & CVPR'21 & 82.1 & 85.7 & 47.0 & 52.3 \\
        BCL \citep{zhu2022balanced} & CVPR'22 & 84.3 & 87.2 & 51.9 & 56.6 \\
        GCL \citep{Li2022GCL} & CVPR'22 & 82.7 & 85.5 & 48.7 & 53.6 \\
        TSC \citep{li2022targeted} & CVPR'22 & 79.7 & 82.9 & 43.8 & 47.4 \\
        ResLT \citep{cui2022reslt} & TPAMI'22 & 82.4 & 85.2 & 48.2 & 52.7 \\
        GLMC\textsuperscript{\textdagger} \citep{du2023global} & CVPR'23 & \textbf{87.8} & 90.3 & 57.5 & 62.5 \\
        MDCS \citep{zhao2023mdcs} & ICCV'23 & 85.8 & 89.4 & 53.2 & 57.2 \\
        GPaCo \citep{cui2023generalized} & TPAMI'23 & - & - & 52.3 & 56.4 \\
        SURE + RW \citep{Li_2024_CVPR} & CVPR'24 & 86.9 & 90.2 & 57.3 & \textbf{63.1} \\
        \midrule
		BoT + ID [\textbf{Ours}] & - & 81.8 & 85.1 & 48.6 & 53.1 \\
        SURE + RW + ID [\textbf{Ours}] & - & \underline{87.0} & \underline{90.4} & \textbf{57.7} & 62.7  \\
        GLMC + ID [\textbf{Ours}] & - & 86.8 & \textbf{90.6} & \underline{57.6} & \underline{62.8} \\
		\bottomrule
	\end{tabular}
  	\caption{\textbf{Experiment 2:} Top-1 accuracy results for state-of-the-art methods 
	on long-tailed CIFAR datasets with a ResNet-32 backbone. \textsuperscript{\textdagger} denotes results reproduced using the code released by the authors.}
    \label{tab:cifarlt-sota}
\end{table*}
\begin{table}[hbt!]
	\centering
    \small
	\begin{tabular}{l r| c }
        \toprule
        \multicolumn{2}{l|}{Method} & Acc. \\
        \midrule\midrule
        OLTR \citep{liu2019large} \hfill & CVPR'19 & 35.9 \\
        BALMS \citep{ren2020balanced} \hfill & NeurIPS'20 & 38.7 \\
        BoT \citep{zhang2021bag} \hfill & AAAI'21 & \underline{43.1} \\
        DisAlign \citep{zhang2021distribution} \hfill & CVPR'21 & 39.5 \\
        SADE \citep{zhang2022self} (3 experts) \hfill\ \ & NeurIPS'22 & 40.9 \\
        ResLT \citep{cui2022reslt} \hfill & TPAMI'22 & 39.8 \\
        MDCS \citep{zhao2023mdcs}  \hfill & ICCV'23 & 42.4 \\
        GPaCo \citep{cui2023generalized} \hfill & TPAMI'23 & 41.7 \\
        APA + AGLU \citep{alexandridis2024adaptive} \hfill & ECCV'24 & 42.0 \\
        \midrule
        PB-ID [\textbf{Ours}] \hfill\    &  - & {42.6 $\pm$ 0.15}        \\
		BoT + ID [\textbf{Ours}]\hfill\                  &  - & {\textbf{43.4 $\pm$ 0.09}} \\
		\bottomrule
	\end{tabular}
	\caption{\textbf{Experiment 2:} Top-1 accuracy results for state-of-the-art methods on Places-LT with a ResNet-152 backbone.}
	\label{table:placeslt-sota}
\end{table} 

\subsection{Experiment 3: ID and Semantic Imbalance}

We now consider a scenario with ``semantic imbalance'', where classes have the same cardinalities but different semantic difficulties. With cardinalities being the same across classes, \textbf{this form of imbalance is impossible to address with cardinality-based imbalance mitigation methods}.
To this end, we use a semantically-imbalanced natural image dataset, SVCI-20 \citep{baltaci2023class}, containing data from 20 classes -- 10 classes each from the CIFAR-10 and SVHN datasets, respectively. The dataset is processed such that all classes have the same number of training samples (by random sampling $N$ examples per class), leaving semantic imbalance as the prominent form of imbalance in the dataset.

\begin{figure}[hbt!]
    \centering
    \includegraphics[width=0.9\linewidth]{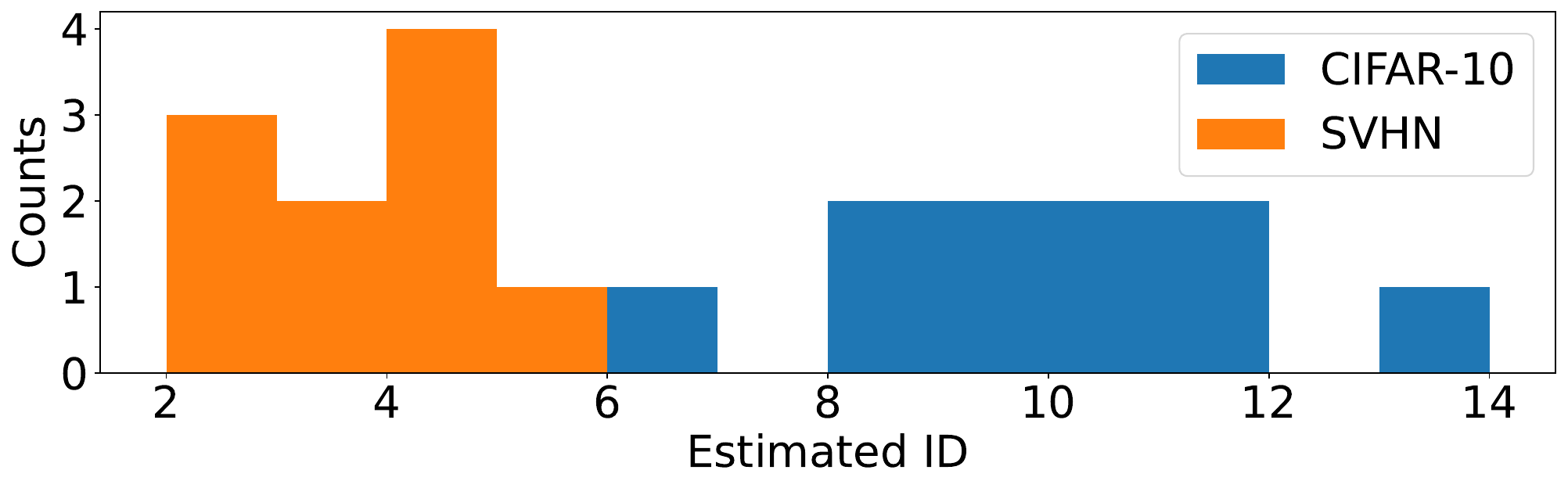}
    \caption{\textbf{Experiment 3:} Histogram of estimated ID values of the two semantically-different types of classes in the SVCI-20 dataset. See Appendix \ref{sect:robustness_analysis} for all ID values.}
    \label{fig:svci-classes-id-comparison}
\end{figure}

As shown in Fig. \ref{fig:svci-classes-id-comparison}, the individual ID estimates of CIFAR-10 classes are much higher than SVHN classes, indicating a clear separation of CIFAR-10 classes from SVHN classes within the SVCI-20 dataset when using ID-based model training. This shows that our ID-based approach does capture imbalance among classes where cardinality-based methods fail. Individual ID values for SVCI-20 classes can additionally be found in Appendix \ref{sect:robustness_analysis}.

Imbalance mitigation results in Table \ref{table:id-semantic-imbalance} indicate that our ID-based approach provides better results and is not affected by cardinality, unlike the cardinality-based approaches.

\noindent\textbf{Summary}. We demonstrate that cardinality-based methods fail in datasets with classes of equal cardinality but different semantic complexity. However, our ID-based approach is effective in such semantic imbalance scenarios.
\begin{table}[hbt!]
   \small
   \centering
   \begin{tabular}{l | c c}
   \toprule
        & $ N = 250$ & $N = 100$ \\
    \midrule\midrule
    CE / Cardinality \citep{cui2019class,kang2019decoupling,mahajan2018exploring,japkowicz2002class}  & 77.96 \scriptsize{$\pm$ 0.65} & 66.44 \scriptsize{$\pm$ 1.40} \\
    Focal Loss \citep{lin2017focal} & 78.35 \scriptsize{$\pm$ 0.04} & 66.39 \scriptsize{$\pm$ 0.97} \\
    \midrule
    PB-ID [\textbf{Ours}] & \textbf{79.02 \scriptsize{$\pm$ 0.18}} & \textbf{67.29 \scriptsize{$\pm$ 0.84}} \\
    \bottomrule
    \end{tabular}
    \caption{\textbf{Experiment 3:} Top-1 balanced accuracy values for plug-and-play resampling and reweighting methods on the balanced SVCI-20 dataset. $N$: Number of samples per class.
    }
    \label{table:id-semantic-imbalance}
\end{table}

\subsection{Experiment 4: Ablation on the Choice of ID Estimator}
Considering that there are many ways to estimate the intrinsic dimensionality of a class, we now evaluate our ID-based approach with different ID estimators: FisherS \citep{albergante2019estimating},  maximum likelihood estimator (MLE) \citep{levina2004maximum} with asymptotic correction  \citep{mackay2005comments}, and Tight Locality Estimation (TLE)  \citep{amsaleg2019intrinsic}.
For comparison on CIFAR-LT, we use the GLMC \citep{du2023global} and SURE + re-weighting methods \citep{Li_2024_CVPR}, and for ImageNet-LT, we use BCL \citep{zhu2022balanced}.
Table \ref{tab:cifarlt-id-ablation} shows the results for our method paired up with the GLMC and SURE on CIFAR-LT datasets, 
while Table~\ref{tab:imagenetlt-id-ablation}
indicates that for ImageNet-LT, the FisherS estimator performs better than alternative estimators when paired up with the BCL \citep{zhu2022balanced} method.

In general, our results indicate that there is no singular optimal choice of ID estimator for all cases, and that the optimal estimator may be different for a particular dataset and a particular mitigation method.
However, based on its all-around performance improvements on multiple datasets and its demonstrated empirical stability for natural image datasets,
we suggest that our ID-based method paired with the FisherS estimator provides the most stable and consistent results.

\noindent\textbf{Summary}. Overall, we see that our approach works with other ID estimators and that there is no single ID estimator which is best for all methods. With SURE + ID and BCL + ID, our method equipped with the FisherS estimator works best, while with GLMC + ID we observe that TLE and MLE provides better results. Considering all options, our method with the FisherS estimator provides the most consistent improvements.

\begin{table*}[hbt!]
	\centering
    \footnotesize
    \begin{tabular}{l r | c c c c}
		\toprule 
    \multicolumn{2}{c}{Category:} & Many    & Med.  & Few  & Total \\
            \midrule\midrule
        \rowcolor{gray!30!}
        \multicolumn{6}{c}{Backbone: ResNet-50, Training: $\leq 100$ epochs} \\
        \midrule
            LDAM \citep{cao2019learning} & NeurIPS'19 & 60.4 & 46.9 & 30.7 & 49.8 \\
            BoT \citep{zhang2021bag} & AAAI'21 & - & - & - & 50.8 \\
            Logit Adj. \citep{menon2021longtail} & ICLR'21 & 61.1 & 47.5 & 27.6 & 50.1 \\
            RIDE \citep{wang2020long} (3E) & ICLR'21 & - & - & - & 54.9 \\
            MiSLAS \citep{zhong2021improving} & CVPR'21 & 62.9 & 50.7 & 34.3 & 52.7 \\
            ACE \citep{cai2021ace} (3E) & ICCV'21 & - & - & - & 54.7 \\
            BCL \citep{zhu2022balanced} & CVPR'22 & - & - & - & \underline{56.0} \\
            GCL \citep{li2022long} & CVPR'22 & 63.0 & \underline{52.7} & \underline{37.1} & 54.5 \\
            TSC \citep{li2022targeted} & CVPR'22 & \underline{63.5} & 49.7 & 30.4 & 52.4 \\
        \midrule
        BoT + ID [\textbf{Ours}] & - & - & - & - & 51.3 \\
        BCL + ID [\textbf{Ours}] (90EP) & - & \textbf{66.6} & \textbf{53.8} & \textbf{37.6} & \textbf{56.5} \\
        \midrule
        \rowcolor{gray!30!}
        \multicolumn{6}{c}{Backbone: ResNeXt-50, Training: $\leq 100$ epochs} \\
        \midrule
        BALMS\textsuperscript{\textdagger} \citep{ren2020balanced} & NeurIPS'20 & 52.2 & 48.8 & 29.8 & 51.4 \\
        RIDE \citep{wang2020long} (3E)&  ICLR'21 & 68.0 & 52.9 & 35.1 & 56.3 \\
        DisAlign \citep{zhang2021distribution} & CVPR'21 & 62.7 & 52.1 & 31.4 & 53.4 \\
        ACE \citep{cai2021ace} (3E) & ICCV'21 & \textbf{71.7} & 54.6 & 23.5 & 56.6 \\
        BCL\textsuperscript{\textdagger} \citep{zhu2022balanced} & CVPR'22 & 67.6 & 54.6 & 37.5 & \underline{57.3} \\
        ResLT \citep{cui2022reslt} & TPAMI'22 & 63.6 & \textbf{55.7} & \textbf{38.9} & 56.1 \\
        \midrule
        BCL\textsuperscript{\textdagger} + ID [\textbf{Ours}] & - & \underline{69.0} & \underline{54.7} & \underline{37.8} & \textbf{57.9} \\
        \midrule
        \rowcolor{gray!30!}
        \multicolumn{6}{c}{Backbone: ResNeXt-50, Training: $\leq 180$ epochs} \\
        \midrule
        BALMS\textsuperscript{\textdagger} \citep{ren2020balanced} & NeurIPS'20 & 65.8 & 53.2 &  34.1 &  55.4 \\
        PaCo\textsuperscript{\textdagger} \citep{cui2021parametric} (180EP) & ICCV'21 & 64.4 & \underline{55.7} & 33.7 & 56.0 \\
        BCL \citep{zhu2022balanced} & CVPR'22 & - & - & - & 57.1 \\
        SADE \citep{zhang2022self} (3E) & NeurIPS'22 & 67.0 & \textbf{56.7} & \textbf{42.6} & \underline{58.8} \\
        GLMC \citep{du2023global} (135EP) & CVPR'23 & \underline{70.1} & 52.4 & 30.4 & 56.3 \\
        DSB + RIDE \citep{ma2022delving} & ICLR'23 & 68.6 & 54.5 & \underline{38.5} & 58.2 \\
        MDCS\textsuperscript{\textdagger} \citep{zhao2023mdcs} (3E)&  ICCV'23 & - & - & - & \textbf{60.2} \\
        \midrule
        GLMC + ID [\textbf{Ours}]& - & 66.8 & 53.7 & 34.4 & 56.3 \\
        BCL\textsuperscript{\textdagger} + ID [\textbf{Ours}] & - & \textbf{70.9} & 53.9 & 37.2 & 58.2 \\
		\bottomrule
	\end{tabular}
	\caption{\textbf{Experiment 2:} Top-1 accuracy results for state-of-the-art methods on ImageNet-LT under 100 and 180 epochs. \textsuperscript{\textdagger} denotes methods that use RandAugment \citep{cubuk2020randaugment} augmentation. 3E: 3 experts. EP: Epochs.}
	\label{tab:imagenetlt-sota}
\end{table*}

\begin{table*}
   \small
   \centering
   \begin{tabular}{l | c c | c c }
		\toprule 
        \multicolumn{1}{c}{Dataset:} & \multicolumn{2}{c}{CIFAR-10-LT} & \multicolumn{2}{c}{CIFAR-100-LT} \\
		\midrule 
        \multicolumn{1}{c}{Imbalance Ratio ($\rho$):}    & 100      & 50        & 100     &  50     \\
        \midrule\midrule
        SURE + RW \citep{Li_2024_CVPR} & \underline{86.9} & \underline{90.2} & \underline{57.3} & \textbf{63.1}  \\
        SURE + RW + ID (w/ MLE) & 86.2 & 90.0 & 54.8 & 59.7 \\
        SURE + RW + ID (w/ TLE) & 86.0 & 90.0 & 56.0 & 60.8  \\
        SURE + RW + ID (w/ FisherS) & \textbf{87.0} & \textbf{90.4} & \textbf{57.7} & \underline{62.7}  \\
        \midrule
        \midrule
        GLMC\textsuperscript{\textdagger} \citep{du2023global} & \underline{87.8} & 90.3 & 57.5 & 62.5  \\
        GLMC + ID (w/ MLE) & 87.1 & \textbf{90.7} & 57.2 & \textbf{63.0}  \\
        GLMC + ID (w/ TLE) & \textbf{87.9} & 90.5 & \textbf{58.0} & \underline{62.8} \\
        GLMC + ID (w/ FisherS) & 86.8 & \underline{90.6} & \underline{57.6} & \underline{62.8}  \\
		\bottomrule
	\end{tabular}
  	\caption{\textbf{Experiment 4:} Top-1 accuracy results for ID-based methods with different ID estimators on CIFAR-LT. \textsuperscript{\textdagger} denotes results reproduced using the code released by the authors.
    }
    \label{tab:cifarlt-id-ablation}
\end{table*}
\begin{table}[hbt!]
   \small
   \centering
   \begin{tabular}{l | c }
		\toprule \multicolumn{1}{c|}{Dataset:} & ImageNet-LT \\
        \midrule
        BCL \citep{zhu2022balanced} & \underline{56.1} \\
        BCL + ID (w/ MLE) & 54.4  \\
        BCL + ID (w/ TLE) & 55.5  \\
        BCL + ID (w/ FisherS) & \textbf{56.5} \\
		\bottomrule
	\end{tabular}
  	\caption{\textbf{Experiment 4:} Top-1 accuracy results for ID-based methods with different ID estimators on ImageNet-LT with a ResNet-50 backbone trained for 90 epochs.}
    \label{tab:imagenetlt-id-ablation}
\end{table}

\subsection{Experiment 5: Failure Cases with ID-based Mitigation}

\label{sect:failure_cases}

We also briefly explore some cases where our ID-based method may fail to be effective in imbalance mitigation.
One possible failure case for our methods arises when the estimated ID fails to be informative for a particular class. One way to analyze this is to reverse or shuffle the estimated ID values for each class in the dataset. Table~\ref{tab:id-failure-cases} shows the result of mitigation with uninformative ID values on CIFAR-10, where the results trail behind standard CE-based training. 

\begin{table}[hbt!]
\small
\centering
\begin{tabular}{l|c|c}
\toprule
    Dataset: &  \multicolumn{1}{c|}{CIFAR-10-LT} & \multicolumn{1}{c}{CIFAR-100-LT} \\ \midrule
CE & 71.60 & 38.90 \\
CSCE \citep{japkowicz2002class} & 74.69 & 41.08 \\
ID & 76.44 & 41.65 \\
\midrule
ID (Reversed) & 71.55 & 39.33 \\
ID (Shuffled) & 71.52 \scriptsize{$\pm$ 0.85} & 38.84 \scriptsize{$\pm$ 0.61} \\
\bottomrule
\end{tabular}
\caption{\textbf{Experiment 5:} Top-1 accuracy results with uninformative ID configurations for CIFAR-LT datasets with an imbalance ratio of 100.}
\label{tab:id-failure-cases}
\end{table}

\begin{figure}[hbt!]
     \centering
   \includegraphics[width=\linewidth]{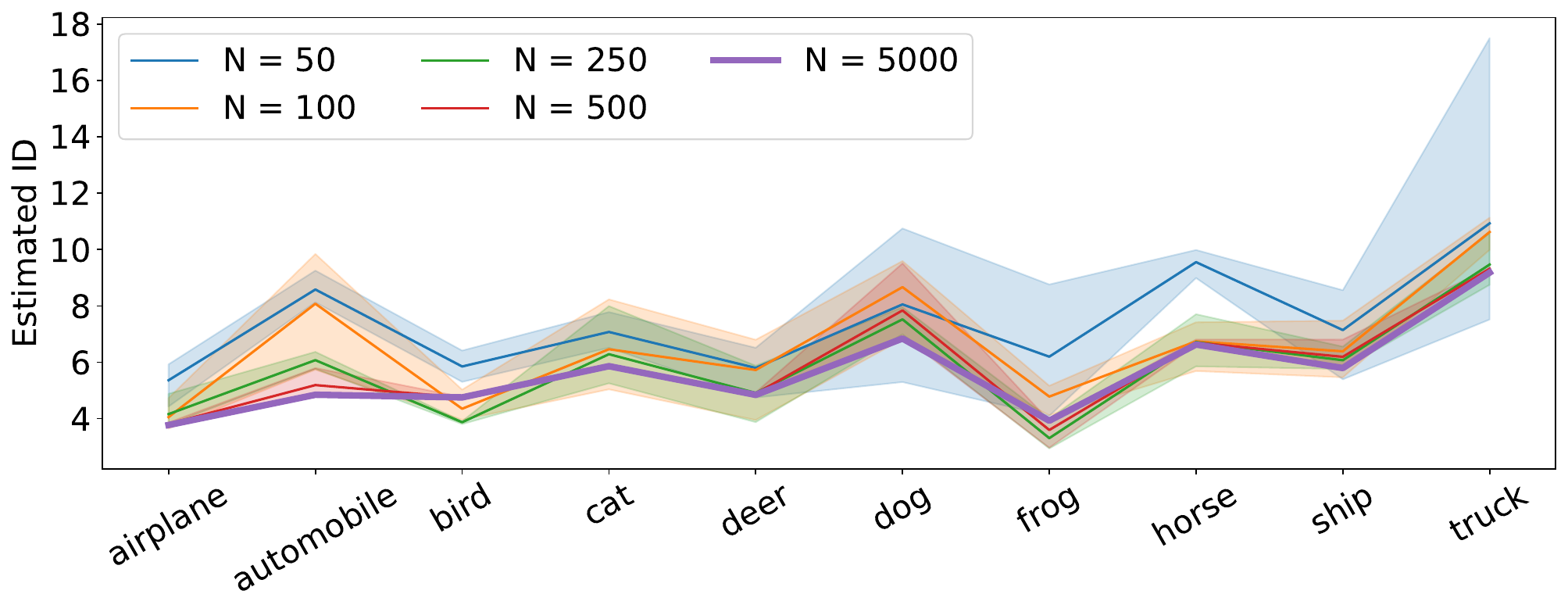}
   \caption{\textbf{Experiment 5:} ID estimates on CIFAR-10 with low sample sizes.}
   \label{fig:id-with-low-sample-size}
\end{figure}

Another possible failure case is in the extreme case where there are not enough samples to reliably estimate ID for a class; for example, estimating ID for a class with just one sample is not feasible. In particular, Fig.~\ref{fig:id-with-low-sample-size} shows that the variance in estimated ID increases with very low sample sizes, which indicates that ID estimates for the tail-most classes in datasets with extreme imbalance might be uninformative due to high variance.

\subsubsection{Fallback Strategies for ID-based Mitigation}
In this section, we briefly consider two fallback strategies for ID-based imbalance mitigation in the case of noisy ID estimates.

\begin{figure}[hbt!]
    \centering
    \includegraphics[width=0.9\linewidth]{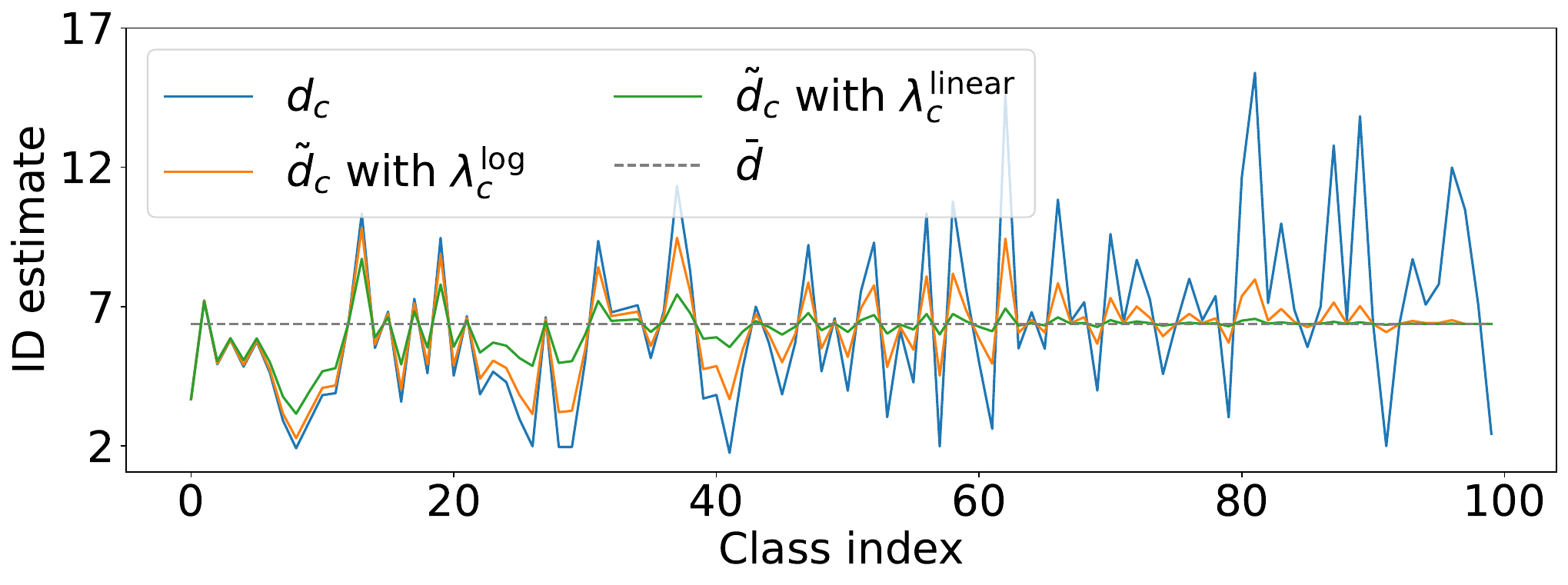}
    \caption{\textbf{Experiment 5:} ID estimates with mean-based fallback strategies for CIFAR-100-LT with an imbalance ratio of $\rho = 50$.}
    \label{fig:id-regression}
\end{figure}

First, as a simpler construction to handle noisy ID estimates, we consider a combination strategy for tail classes in the dataset in the following manner:
We modify the class-wise ID estimates $d_c$ as $\tilde{d_c}$, which is defined as:
\begin{equation}
    \tilde{d_c} = \lambda_c d_c + (1-\lambda_c)\bar{d},
\end{equation}
where $d_c$ is the estimated ID for a class,
$\bar{d} = \frac{1}{\mathcal{|C|}}\sum_{c}d_c$ represents the mean ID of all classes in the dataset, and
$\lambda_c$ is a scaling parameter that scales the weighting depending on the cardinality of the class $c$. We consider two schedules for $\lambda_c$, a linear scaling which we define as
\begin{equation}
    \lambda_c^{\mathrm{linear}} = \frac{N_c - N_{\min}}{N_{\max} - N_{\min}},
\end{equation}
and a log-scaled version defined as
\begin{equation}
\lambda_c^{\mathrm{log}} = \frac{\log N_c - \log N_{\min}}{\log N_{\max} - \log N_{\min}}.
\end{equation}
This mean-based formulation of $\tilde{d_c}$ reduces the variance of ID estimates for tail classes by moving $\tilde{d_c}$ towards the mean ID.
Fig.~\ref{fig:id-regression} shows an example of combined ID values for CIFAR-100-LT.

\begin{figure}[hbt!]
    \centering
    \includegraphics[width=0.9\linewidth]{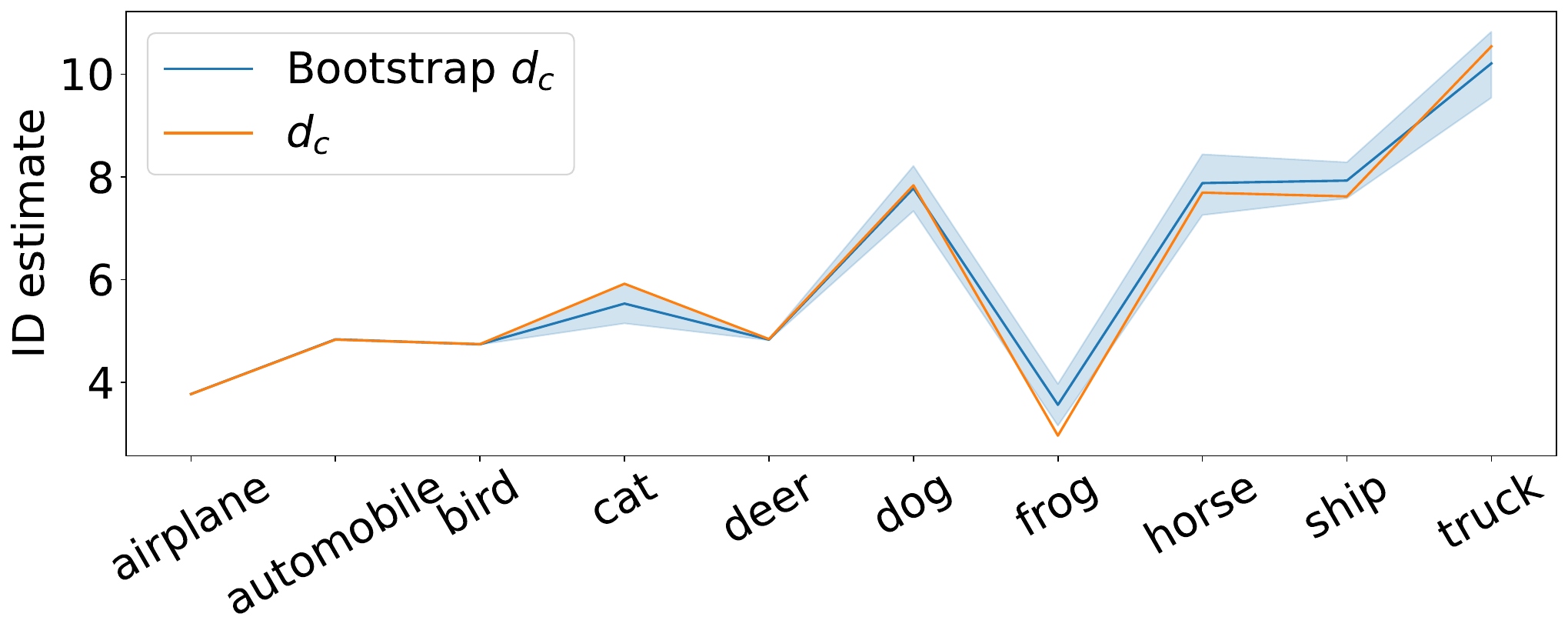}
    \caption{\textbf{Experiment 5:} ID class-wise estimates ($d_c$) along with their bootstrapped ID ranges, for CIFAR-10-LT with an imbalance ratio of $\rho = 50$.}
    \label{fig:id-bootstrap}
\end{figure}

Second, we adopt a gating-based approach based on the following strategy: for each class, we construct a bootstrap distribution from repeated ID estimates with random 80\% subsampling of their original samples.
We then calculate confidence intervals for the ID estimates of each class, and gate our mitigation strategy with the following rule: if the CI range of the estimated ID is over a fixed value, we replace the ID-based $d_c$ weighting for that class with the cardinality-based weighting \cite{japkowicz2002class} instead. 
Fig.~\ref{fig:id-bootstrap} shows an example of bootstrapped ID for CIFAR-10-LT with an imbalance ratio of $\rho = 100$.

Table~\ref{tab:id-fallback-table} shows the results of training model on CIFAR-LT datasets with these two fallback strategies. In particular, while ID with cardinality-based gating does not seem to have much effect, ID with mean fallback seems to slightly improve performance for the CIFAR-100-LT dataset with an imbalance ratio of 100.
We hypothesize that gating might not be effective due to the mixing of ID-based and cardinality-based weights, while mean fallback may be useful for datasets with many tails of low sample sizes.
\begin{table*}[hbt!]
    \centering
    \small
    \begin{tabular}{l|c c c c}
    \toprule
    \multicolumn{1}{c|}{Dataset:} & \multicolumn{2}{c}{CIFAR-10-LT} & \multicolumn{2}{c}{CIFAR-100-LT} \\
    \midrule
    \multicolumn{1}{c|}{Imbalance Ratio ($\rho$):} & 100 & 50 & 100 & 50 \\
    \midrule
    \midrule
    ID & \textbf{76.4} & 80.8 & 41.7 & \textbf{46.2} \\
    \midrule
    ID w/ mean fallback ($\lambda_{c}^{\mathrm{linear}}$) & 75.9 \scriptsize{$\pm$ 0.7} & \textbf{80.9 \scriptsize{$\pm$ 0.3}} & \textbf{42.1 \scriptsize{$\pm$ 0.2}} & 45.9 \scriptsize{$\pm$ 0.3} \\
    ID w/ mean fallback ($\lambda_{c}^{\mathrm{log}}$) & 75.8 \scriptsize{$\pm$ 0.5} & 80.8 \scriptsize{$\pm$ 0.4} & 41.5 \scriptsize{$\pm$ 0.4} & 45.8 \scriptsize{$\pm$ 0.5} \\
    ID w/ gating fallback & 75.3 \scriptsize{$\pm$ 0.5} & 80.5 \scriptsize{$\pm$ 0.4} & 40.2 \scriptsize{$\pm$ 0.3} & 45.8 \scriptsize{$\pm$ 0.6} \\
    \bottomrule
    \end{tabular}
    \caption{\textbf{Experiment 5:} Top-1 accuracy results for fallback strategies for ID-based re-weighting.}
    \label{tab:id-fallback-table}
\end{table*}

{
\noindent\textbf{Summary}. Similar to other mitigation methods in the literature, ID-based mitigation also includes possible failure cases, particularly in the case of classes with extremely low sample sizes which can result in noisy ID estimates. Using uninformative or high-variance ID estimates can shift the model towards focusing on the wrong classes, and lead to suboptimal training and performance.
}
\section{Conclusion and Discussion}
\label{sec:conclusion}

\subsection{Conclusion}
Prior work addresses imbalance by relying on class cardinality or by training multi-stage, complex models.
In this paper, we proposed ``data intrinsic dimensionality (ID)'' as an alternative measure for imbalance.
We show that data ID can measure imbalance in a model-free, training-free, and cardinality-independent way, making it highly practical and versatile for a range of imbalance mitigation methods.
Our experiments show that data ID achieves results that are on par with or better than state-of-the-art methods across several datasets with varying degrees of cardinality imbalance.

{
\subsection{Practical Guidance}
Considering all experiments about ID-based mitigation throughout this section, we suggest the following setup for ID-based mitigation.
\\
\textbf{Estimating ID}.
Considering all available choices for ID estimator, we suggest using the FisherS \cite{albergante2019estimating} estimator with hyperparameter $C = 10$ given its robustness with natural image data. Our experiments indicate that ID estimation based on Fisher separability of data works well on both synthetic and real-life datasets, and is more reliable under higher degrees of imbalance.
}\\
{
\textbf{Using Class-wise ID}.
Our results suggest that ID-based mitigation works best when used as a resampling or a reweighting strategy. Moreover, although our results show that ID estimation is robust to noise in the data, it can be negatively affected by the number of samples in a class. As a fallback option in such cases, we suggest using the linear-scaling mean ID fallback introduced in Section \ref{sect:failure_cases}.
}

{
\subsection{Limitations and Future Work}
ID-based mitigation shows promise with long-tailed datasets, however, accurately estimating the ID of samples with very low sample sizes remains a challenge. We believe that with better ID estimation tools in the future, our method can improved in this limiting aspect.
}

{
Another possible direction in future work regarding ID-based mitigation is to further test the robustness of our proposed method through even broader evaluation across diverse datasets, different modalities and real-world settings.
Similarly, exploring extensions that improve the interpretability of ID-based mitigation or reduce the computation cost is a promising idea to increase the practicality of our proposed approach.
}

\section*{Acknowledgments}
This work was supported by the Council of Higher Education Research Universities Support program as a Middle East Technical University (METU) Scientific Research Projects project (``New Techniques in Visual Recognition’’, Project No. ADEP-312-2024-11485). We also gratefully acknowledge the computational resources provided by METU-ROMER, Center for Robotics and Artificial Intelligence.

\FloatBarrier
\bibliographystyle{unsrtnat}
\bibliography{main}  %

@String(IJCV = {Int. J. Comput. Vis.})

@String(CVPR= {IEEE Conf. Comput. Vis. Pattern Recog.})

@String(ICCV= {Int. Conf. Comput. Vis.})

@String(ECCV= {Eur. Conf. Comput. Vis.})

@String(ICLR = {Int. Conf. Learn. Represent.})

@String(AAAI = {AAAI})

@String(IJCV  = {IJCV})

@String(CVPR  = {CVPR})

@String(ICCV  = {ICCV})

@String(ECCV  = {ECCV})

@String(ICLR  = {ICLR})

@inproceedings{narayanan2009sample,
  title={On the Sample Complexity of Learning Smooth Cuts on a Manifold.},
  author={Narayanan, Hariharan and Niyogi, Partha},
  booktitle={Annual Conference on Learning Theory},
  year={2009}
}

@article{vapnik1999overview,
  title={An overview of statistical learning theory},
  author={Vapnik, Vladimir N},
  journal={IEEE transactions on neural networks},
  volume={10},
  number={5},
  pages={988--999},
  year={1999},
  publisher={IEEE}
}

@article{Shi2023HowRH,
  title={How Re-sampling Helps for Long-Tail Learning?},
  author={Jiang-Xin Shi and Tong Wei and Yuke Xiang and Yu-Feng Li},
  journal={Advances in Neural Information Processing Systems},
  volume={36},
  year={2023}
}

@inproceedings{peng2020large,
  title={Large-scale object detection in the wild from imbalanced multi-labels},
  author={Peng, Junran and Bu, Xingyuan and Sun, Ming and Zhang, Zhaoxiang and Tan, Tieniu and Yan, Junjie},
  booktitle={Proceedings of the IEEE/CVF conference on computer vision and pattern recognition},
  pages={9709--9718},
  year={2020}
}

@InProceedings{taoLGLA2023,
    author    = {Tao, Yingfan and Sun, Jingna and Yang, Hao and Chen, Li and Wang, Xu and Yang, Wenming and Du, Daniel and Zheng, Min},
    title     = {Local and Global Logit Adjustments for Long-Tailed Learning},
    booktitle = {Proceedings of the IEEE/CVF International Conference on Computer Vision (ICCV)},
    month     = {October},
    year      = {2023},
    pages     = {11783-11792}
}

@inproceedings{menon2021longtail,
  author       = {Aditya Krishna Menon and
                  Sadeep Jayasumana and
                  Ankit Singh Rawat and
                  Himanshu Jain and
                  Andreas Veit and
                  Sanjiv Kumar},
  title        = {Long-tail learning via logit adjustment},
  booktitle    = {9th International Conference on Learning Representations, {ICLR} 2021,
                  Virtual Event, Austria, May 3-7, 2021},
  publisher    = {OpenReview.net},
  year         = {2021},
  url          = {https://openreview.net/forum?id=37nvvqkCo5},
  bibsource    = {dblp computer science bibliography, https://dblp.org}
}

@article{fernando2022dynamical,
  author={Fernando, K. Ruwani M. and Tsokos, Chris P.},
  journal={IEEE Transactions on Neural Networks and Learning Systems}, 
  title={Dynamically Weighted Balanced Loss: Class Imbalanced Learning and Confidence Calibration of Deep Neural Networks}, 
  year={2022},
  volume={33},
  number={7},
  pages={2940-2951},
  doi={10.1109/TNNLS.2020.3047335}
}

@inproceedings{Park2021ibloss,
    author    = {Park, Seulki and Lim, Jongin and Jeon, Younghan and Choi, Jin Young},
    title     = {Influence-Balanced Loss for Imbalanced Visual Classification},
    booktitle = {Proceedings of the IEEE/CVF International Conference on Computer Vision (ICCV)},
    month     = {October},
    year      = {2021},
    pages     = {735-744}
}

@inproceedings{jamal2020rethinking,
  title={Rethinking class-balanced methods for long-tailed visual recognition from a domain adaptation perspective},
  author={Jamal, Muhammad Abdullah and Brown, Matthew and Yang, Ming-Hsuan and Wang, Liqiang and Gong, Boqing},
  booktitle={Proceedings of the IEEE/CVF conference on computer vision and pattern recognition},
  pages={7610--7619},
  year={2020}
}

@article{shu2019meta,
  title={Meta-weight-net: Learning an explicit mapping for sample weighting},
  author={Shu, Jun and Xie, Qi and Yi, Lixuan and Zhao, Qian and Zhou, Sanping and Xu, Zongben and Meng, Deyu},
  journal={Advances in neural information processing systems},
  volume={32},
  year={2019}
}

@inproceedings{ren2018learning,
  title={Learning to reweight examples for robust deep learning},
  author={Ren, Mengye and Zeng, Wenyuan and Yang, Bin and Urtasun, Raquel},
  booktitle={International conference on machine learning},
  pages={4334--4343},
  year={2018},
  organization={PMLR}
}

@book{cook1982residuals,
  author    = {R. Dennis Cook and Sanford Weisberg},
  title     = {Residuals and Influence in Regression},
  year      = {1982}
}

@inproceedings{Li2022GCL,
  author    = {Mengke Li, Yiu{-}ming Cheung, Yang Lu},
  title     = {Long-tailed Visual Recognition via Gaussian Clouded Logit Adjustment},
  pages = {6929-6938},
  booktitle = {CVPR},
  year      = {2022},
}

@inproceedings{hong2021disentangling,
  title={Disentangling label distribution for long-tailed visual recognition},
  author={Hong, Youngkyu and Han, Seungju and Choi, Kwanghee and Seo, Seokjun and Kim, Beomsu and Chang, Buru},
  booktitle={Proceedings of the IEEE/CVF conference on computer vision and pattern recognition},
  pages={6626--6636},
  year={2021}
}

@article{kang2019decoupling,
  title={Decoupling representation and classifier for long-tailed recognition},
  author={Kang, Bingyi and Xie, Saining and Rohrbach, Marcus and Yan, Zhicheng and Gordo, Albert and Feng, Jiashi and Kalantidis, Yannis},
  journal={arXiv preprint arXiv:1910.09217},
  year={2019}
}

@article{yang2022survey,
  title={A survey on long-tailed visual recognition},
  author={Yang, Lu and Jiang, He and Song, Qing and Guo, Jun},
  journal={International Journal of Computer Vision},
  volume={130},
  number={7},
  pages={1837--1872},
  year={2022},
  publisher={Springer}
}

@inproceedings{cui2019class,
  title={Class-balanced loss based on effective number of samples},
  author={Cui, Yin and Jia, Menglin and Lin, Tsung-Yi and Song, Yang and Belongie, Serge},
  booktitle={Proceedings of the IEEE/CVF conference on computer vision and pattern recognition},
  pages={9268--9277},
  year={2019}
}

@article{cao2019learning,
  title={Learning imbalanced datasets with label-distribution-aware margin loss},
  author={Cao, Kaidi and Wei, Colin and Gaidon, Adrien and Arechiga, Nikos and Ma, Tengyu},
  journal={Advances in neural information processing systems},
  volume={32},
  year={2019}
}

@inproceedings{liu2019large,
  title={Large-scale long-tailed recognition in an open world},
  author={Liu, Ziwei and Miao, Zhongqi and Zhan, Xiaohang and Wang, Jiayun and Gong, Boqing and Yu, Stella X},
  booktitle={Proceedings of the IEEE/CVF conference on computer vision and pattern recognition},
  pages={2537--2546},
  year={2019}
}

@article{zhang2023deep,
  title={Deep long-tailed learning: A survey},
  author={Zhang, Yifan and Kang, Bingyi and Hooi, Bryan and Yan, Shuicheng and Feng, Jiashi},
  journal={IEEE Transactions on Pattern Analysis and Machine Intelligence},
  volume={45},
  number={9},
  pages={10795--10816},
  year={2023},
  publisher={IEEE}
}

@inproceedings{drummond2003c4,
  title={C4. 5, class imbalance, and cost sensitivity: why under-sampling beats over-sampling},
  author={Drummond, Chris and Holte, Robert C and others},
  booktitle={Workshop on learning from imbalanced datasets II},
  volume={11},
  pages={1--8},
  year={2003}
}

@inproceedings{zhang2021bag,
  title={Bag of tricks for long-tailed visual recognition with deep convolutional neural networks},
  author={Zhang, Yongshun and Wei, Xiu-Shen and Zhou, Boyan and Wu, Jianxin},
  booktitle={Proceedings of the AAAI conference on artificial intelligence},
  volume={35},
  pages={3447--3455},
  year={2021}
}

@book{fukunaga2013introduction,
  title={Introduction to statistical pattern recognition},
  author={Fukunaga, Keinosuke},
  year={2013},
  publisher={Elsevier}
}

@article{camastra2016intrinsic,
  title={Intrinsic dimension estimation: Advances and open problems},
  author={Camastra, Francesco and Staiano, Antonino},
  journal={Information Sciences},
  volume={328},
  pages={26--41},
  year={2016},
  publisher={Elsevier}
}

@inproceedings{albergante2019estimating,
  title={Estimating the effective dimension of large biological datasets using Fisher separability analysis},
  author={Albergante, Luca and Bac, Jonathan and Zinovyev, Andrei},
  booktitle={2019 International Joint Conference on Neural Networks (IJCNN)},
  pages={1--8},
  year={2019},
  organization={IEEE}
}

@inproceedings{lin2017focal,
  title={Focal loss for dense object detection},
  author={Lin, Tsung-Yi and Goyal, Priya and Girshick, Ross and He, Kaiming and Doll{\'a}r, Piotr},
  booktitle={Proceedings of the IEEE international conference on computer vision},
  pages={2980--2988},
  year={2017}
}

@inproceedings{samuel2021distributional,
  title={Distributional robustness loss for long-tail learning},
  author={Samuel, Dvir and Chechik, Gal},
  booktitle={Proceedings of the IEEE/CVF International Conference on Computer Vision},
  pages={9495--9504},
  year={2021}
}

@article{ma2024unveiling,
  title={Unveiling and Mitigating Generalized Biases of DNNs through the Intrinsic Dimensions of Perceptual Manifolds},
  author={Ma, Yanbiao and Jiao, Licheng and Liu, Fang and Li, Lingling and Ma, Wenping and Yang, Shuyuan and Liu, Xu and Chen, Puhua},
  journal={arXiv preprint arXiv:2404.13859},
  year={2024}
}

@article{tenenbaum2000global,
  title={A global geometric framework for nonlinear dimensionality reduction},
  author={Tenenbaum, Joshua B and Silva, Vin de and Langford, John C},
  journal={science},
  volume={290},
  number={5500},
  pages={2319--2323},
  year={2000},
  publisher={American Association for the Advancement of Science}
}

@article{roweis2000nonlinear,
  title={Nonlinear dimensionality reduction by locally linear embedding},
  author={Roweis, Sam T and Saul, Lawrence K},
  journal={science},
  volume={290},
  number={5500},
  pages={2323--2326},
  year={2000},
  publisher={American Association for the Advancement of Science}
}

@article{abdi2010principal,
  title={Principal component analysis},
  author={Abdi, Herv{\'e} and Williams, Lynne J},
  journal={Wiley interdisciplinary reviews: computational statistics},
  volume={2},
  number={4},
  pages={433--459},
  year={2010},
  publisher={Wiley Online Library}
}

@article{fukunaga1971algorithm,
  title={An algorithm for finding intrinsic dimensionality of data},
  author={Fukunaga, Keinosuke and Olsen, David R},
  journal={IEEE Transactions on computers},
  volume={100},
  number={2},
  pages={176--183},
  year={1971},
  publisher={IEEE}
}

@article{bishop1998bayesian,
  title={Bayesian pca},
  author={Bishop, Christopher},
  journal={Advances in neural information processing systems},
  volume={11},
  year={1998}
}

@article{ceruti2014danco,
  title={DANCo: An intrinsic dimensionality estimator exploiting angle and norm concentration},
  author={Ceruti, Claudio and Bassis, Simone and Rozza, Alessandro and Lombardi, Gabriele and Casiraghi, Elena and Campadelli, Paola},
  journal={Pattern recognition},
  volume={47},
  number={8},
  pages={2569--2581},
  year={2014},
  publisher={Elsevier}
}

@article{ansuini2019intrinsic,
  title={Intrinsic dimension of data representations in deep neural networks},
  author={Ansuini, Alessio and Laio, Alessandro and Macke, Jakob H and Zoccolan, Davide},
  journal={Advances in Neural Information Processing Systems},
  volume={32},
  year={2019}
}

@article{levina2004maximum,
  title={Maximum likelihood estimation of intrinsic dimension},
  author={Levina, Elizaveta and Bickel, Peter},
  journal={Advances in neural information processing systems},
  volume={17},
  year={2004}
}

@article{facco2017estimating,
  title={Estimating the intrinsic dimension of datasets by a minimal neighborhood information},
  author={Facco, Elena and d’Errico, Maria and Rodriguez, Alex and Laio, Alessandro},
  journal={Scientific reports},
  volume={7},
  number={1},
  pages={12140},
  year={2017},
  publisher={Nature Publishing Group UK London}
}

@article{camastra2002estimating,
  title={Estimating the intrinsic dimension of data with a fractal-based method},
  author={Camastra, Francesco and Vinciarelli, Alessandro},
  journal={IEEE Transactions on pattern analysis and machine intelligence},
  volume={24},
  number={10},
  pages={1404--1407},
  year={2002},
  publisher={IEEE}
}

@article{japkowicz2002class,
  title={The class imbalance problem: A systematic study},
  author={Japkowicz, Nathalie and Stephen, Shaju},
  journal={Intelligent data analysis},
  volume={6},
  number={5},
  pages={429--449},
  year={2002},
  publisher={IOS Press}
}

@article{ILSVRC15,
Author = {Olga Russakovsky and Jia Deng and Hao Su and Jonathan Krause and Sanjeev Satheesh and Sean Ma and Zhiheng Huang and Andrej Karpathy and Aditya Khosla and Michael Bernstein and Alexander C. Berg and Li Fei-Fei},
Title = {{ImageNet Large Scale Visual Recognition Challenge}},
Year = {2015},
journal   = {International Journal of Computer Vision (IJCV)},
doi = {10.1007/s11263-015-0816-y},
volume={115},
number={3},
pages={211-252}
}

@article{zhou2017places,
  title={Places: A 10 million Image Database for Scene Recognition},
  author={Zhou, Bolei and Lapedriza, Agata and Khosla, Aditya and Oliva, Aude and Torralba, Antonio},
  journal={IEEE Transactions on Pattern Analysis and Machine Intelligence},
  year={2017},
  publisher={IEEE}
}

@article{baltaci2023class,
  title={Class Uncertainty: A Measure to Mitigate Class Imbalance},
  author={Baltaci, Zeynep Sonat and Oksuz, Kemal and Kuzucu, Selim and Tezoren, Kivanc and Konar, Berkin Kerim and Ozkan, Alpay and Akbas, Emre and Kalkan, Sinan},
  journal={arXiv preprint arXiv:2311.14090},
  year={2023}
}

@inproceedings{mahajan2018exploring,
  title={Exploring the limits of weakly supervised pretraining},
  author={Mahajan, Dhruv and Girshick, Ross and Ramanathan, Vignesh and He, Kaiming and Paluri, Manohar and Li, Yixuan and Bharambe, Ashwin and Van Der Maaten, Laurens},
  booktitle={Proceedings of the European conference on computer vision (ECCV)},
  pages={181--196},
  year={2018}
}

@article{gorban2018correction,
  title={Correction of AI systems by linear discriminants: Probabilistic foundations},
  author={Gorban, Alexander N and Golubkov, Alexander and Grechuk, Bogdan and Mirkes, Eugenij Moiseevich and Tyukin, Ivan Yu},
  journal={Information Sciences},
  volume={466},
  pages={303--322},
  year={2018},
  publisher={Elsevier}
}

@article{ren2020balanced,
  title={Balanced meta-softmax for long-tailed visual recognition},
  author={Ren, Jiawei and Yu, Cunjun and Ma, Xiao and Zhao, Haiyu and Yi, Shuai and others},
  journal={Advances in neural information processing systems},
  volume={33},
  pages={4175--4186},
  year={2020}
}

@inproceedings{zhong2021improving,
  title={Improving calibration for long-tailed recognition},
  author={Zhong, Zhisheng and Cui, Jiequan and Liu, Shu and Jia, Jiaya},
  booktitle={Proceedings of the IEEE/CVF conference on computer vision and pattern recognition},
  pages={16489--16498},
  year={2021}
}

@article{cui2023generalized,
  title={Generalized parametric contrastive learning},
  author={Cui, Jiequan and Zhong, Zhisheng and Tian, Zhuotao and Liu, Shu and Yu, Bei and Jia, Jiaya},
  journal={IEEE Transactions on Pattern Analysis and Machine Intelligence},
  year={2023},
  publisher={IEEE}
}

@article{cui2022reslt,
  title={Reslt: Residual learning for long-tailed recognition},
  author={Cui, Jiequan and Liu, Shu and Tian, Zhuotao and Zhong, Zhisheng and Jia, Jiaya},
  journal={IEEE transactions on pattern analysis and machine intelligence},
  volume={45},
  number={3},
  pages={3695--3706},
  year={2022},
  publisher={IEEE}
}

@inproceedings{zhang2021distribution,
  title={Distribution alignment: A unified framework for long-tail visual recognition},
  author={Zhang, Songyang and Li, Zeming and Yan, Shipeng and He, Xuming and Sun, Jian},
  booktitle={Proceedings of the IEEE/CVF conference on computer vision and pattern recognition},
  pages={2361--2370},
  year={2021}
}

@inproceedings{li2022targeted,
  title={Targeted supervised contrastive learning for long-tailed recognition},
  author={Li, Tianhong and Cao, Peng and Yuan, Yuan and Fan, Lijie and Yang, Yuzhe and Feris, Rogerio S and Indyk, Piotr and Katabi, Dina},
  booktitle={Proceedings of the IEEE/CVF conference on computer vision and pattern recognition},
  pages={6918--6928},
  year={2022}
}

@inproceedings{zhou2020bbn,
  title={Bbn: Bilateral-branch network with cumulative learning for long-tailed visual recognition},
  author={Zhou, Boyan and Cui, Quan and Wei, Xiu-Shen and Chen, Zhao-Min},
  booktitle={Proceedings of the IEEE/CVF conference on computer vision and pattern recognition},
  pages={9719--9728},
  year={2020}
}

@inproceedings{li2022long,
  title={Long-tailed visual recognition via gaussian clouded logit adjustment},
  author={Li, Mengke and Cheung, Yiu-ming and Lu, Yang},
  booktitle={Proceedings of the IEEE/CVF Conference on Computer Vision and Pattern Recognition},
  pages={6929--6938},
  year={2022}
}

@inproceedings{wang2020long,
  title={Long-tailed Recognition by Routing Diverse Distribution-Aware Experts},
  author={Wang, Xudong and Lian, Long and Miao, Zhongqi and Liu, Ziwei and Yu, Stella},
  booktitle={International Conference on Learning Representations},
  year={2021}
}

@inproceedings{zhu2022balanced,
  title={Balanced contrastive learning for long-tailed visual recognition},
  author={Zhu, Jianggang and Wang, Zheng and Chen, Jingjing and Chen, Yi-Ping Phoebe and Jiang, Yu-Gang},
  booktitle={Proceedings of the IEEE/CVF Conference on Computer Vision and Pattern Recognition},
  pages={6908--6917},
  year={2022}
}

@inproceedings{du2023global,
  title={Global and local mixture consistency cumulative learning for long-tailed visual recognitions},
  author={Du, Fei and Yang, Peng and Jia, Qi and Nan, Fengtao and Chen, Xiaoting and Yang, Yun},
  booktitle={Proceedings of the IEEE/CVF Conference on Computer Vision and Pattern Recognition},
  pages={15814--15823},
  year={2023}
}

@InProceedings{Li_2024_CVPR,
    author    = {Li, Yuting and Chen, Yingyi and Yu, Xuanlong and Chen, Dexiong and Shen, Xi},
    title     = {SURE: SUrvey REcipes for building reliable and robust deep networks},
    booktitle = {Proceedings of the IEEE/CVF Conference on Computer Vision and Pattern Recognition (CVPR)},
    month     = {June},
    year      = {2024},
    pages     = {17500-17510}
}

@article{alexandridis2024adaptive,
  title={Adaptive Parametric Activation},
  author={Alexandridis, Konstantinos Panagiotis and Deng, Jiankang and Nguyen, Anh and Luo, Shan},
  journal={arXiv preprint arXiv:2407.08567},
  year={2024}
}

@inproceedings{amsaleg2019intrinsic,
  title={Intrinsic dimensionality estimation within tight localities},
  author={Amsaleg, Laurent and Chelly, Oussama and Houle, Michael E and Kawarabayashi, Ken-ichi and Radovanovi{\'c}, Milo{\v{s}} and Treeratanajaru, Weeris},
  booktitle={Proceedings of the 2019 SIAM international conference on data mining},
  pages={181--189},
  year={2019},
  organization={SIAM}
}

@inproceedings{he2016deep,
  title={Deep residual learning for image recognition},
  author={He, Kaiming and Zhang, Xiangyu and Ren, Shaoqing and Sun, Jian},
  booktitle={Proceedings of the IEEE conference on computer vision and pattern recognition},
  pages={770--778},
  year={2016}
}

@article{zhang2017mixup,
  title={mixup: Beyond empirical risk minimization},
  author={Zhang, Hongyi},
  journal={arXiv preprint arXiv:1710.09412},
  year={2017}
}

@inproceedings{xie2017aggregated,
  title={Aggregated residual transformations for deep neural networks},
  author={Xie, Saining and Girshick, Ross and Doll{\'a}r, Piotr and Tu, Zhuowen and He, Kaiming},
  booktitle={Proceedings of the IEEE conference on computer vision and pattern recognition},
  pages={1492--1500},
  year={2017}
}

@article{campadelli2015intrinsic,
  title={Intrinsic dimension estimation: Relevant techniques and a benchmark framework},
  author={Campadelli, Paola and Casiraghi, Elena and Ceruti, Claudio and Rozza, Alessandro},
  journal={Mathematical Problems in Engineering},
  volume={2015},
  number={1},
  pages={759567},
  year={2015},
  publisher={Wiley Online Library}
}

@inproceedings{zhao2023mdcs,
  title={Mdcs: More diverse experts with consistency self-distillation for long-tailed recognition},
  author={Zhao, Qihao and Jiang, Chen and Hu, Wei and Zhang, Fan and Liu, Jun},
  booktitle={Proceedings of the IEEE/CVF International Conference on Computer Vision},
  pages={11597--11608},
  year={2023}
}

@inproceedings{cai2021ace,
  title={Ace: Ally complementary experts for solving long-tailed recognition in one-shot},
  author={Cai, Jiarui and Wang, Yizhou and Hwang, Jenq-Neng},
  booktitle={Proceedings of the IEEE/CVF international conference on computer vision},
  pages={112--121},
  year={2021}
}

@inproceedings{cui2021parametric,
  title={Parametric contrastive learning},
  author={Cui, Jiequan and Zhong, Zhisheng and Liu, Shu and Yu, Bei and Jia, Jiaya},
  booktitle={Proceedings of the IEEE/CVF international conference on computer vision},
  pages={715--724},
  year={2021}
}

@article{87e6ea2d996a4ec09e00c4a4c4e9b9a0,
title = "Collinearity: a review of methods to deal with it and a simulation study evaluating their performance : open access",
author = "C.F. Dormann and J. Elith and S. Bacher and G.C.G. Carr{\'e} and {Garc{\'i}a M{\'a}rquez}, J.R. and B. Gruber and B. Lafourcade and P.J. Leitao and T. M{\"u}nkem{\"u}ller and C.J. McClean and P.E. Osborne and B. Reneking and B. Schr{\"o}der and A.K. Skidmore and D. Zurell and S. Lautenbach",
year = "2013",
doi = "10.1111/j.1600-0587.2012.07348.x",
language = "English",
volume = "36",
pages = "27--46",
journal = "Ecography",
issn = "0906-7590",
publisher = "Wiley-Blackwell",
number = "1",

}

@inproceedings{hein2005intrinsic,
  title={Intrinsic dimensionality estimation of submanifolds in Rd},
  author={Hein, Matthias and Audibert, Jean-Yves},
  booktitle={Proceedings of the 22nd international conference on Machine learning},
  pages={289--296},
  year={2005}
}

@mastersthesis{vonLindheim2018,
  author      = {Johannes von Lindheim},
  title       = {On Intrinsic Dimension Estimation and Minimal Diffusion Maps Embeddings of Point Clouds},
  pages       = {61},
  year        = {2018},
  school      = {Freien Universitat Berlin}
}

@article{chawla2002smote,
  title={SMOTE: synthetic minority over-sampling technique},
  author={Chawla, Nitesh V and Bowyer, Kevin W and Hall, Lawrence O and Kegelmeyer, W Philip},
  journal={Journal of artificial intelligence research},
  volume={16},
  pages={321--357},
  year={2002}
}

@article{mackay2005comments,
  title={Comments on ‘Maximum likelihood estimation of intrinsic dimension’by E},
  author={MacKay, David JC and Ghahramani, Zoubin},
  journal={Levina and P. Bickel},
  year={2005}
}

@article{zhang2022self,
  title={Self-supervised aggregation of diverse experts for test-agnostic long-tailed recognition},
  author={Zhang, Yifan and Hooi, Bryan and Hong, Lanqing and Feng, Jiashi},
  journal={Advances in Neural Information Processing Systems},
  volume={35},
  pages={34077--34090},
  year={2022}
}

@inproceedings{cubuk2020randaugment,
  title={Randaugment: Practical automated data augmentation with a reduced search space},
  author={Cubuk, Ekin D and Zoph, Barret and Shlens, Jonathon and Le, Quoc V},
  booktitle={Proceedings of the IEEE/CVF conference on computer vision and pattern recognition workshops},
  pages={702--703},
  year={2020}
}

@article{fefferman2016testing,
  title={Testing the manifold hypothesis},
  author={Fefferman, Charles and Mitter, Sanjoy and Narayanan, Hariharan},
  journal={Journal of the American Mathematical Society},
  volume={29},
  number={4},
  pages={983--1049},
  year={2016}
}

@article{pope2021intrinsic,
  title={The intrinsic dimension of images and its impact on learning},
  author={Pope, Phillip and Zhu, Chen and Abdelkader, Ahmed and Goldblum, Micah and Goldstein, Tom},
  journal={arXiv preprint arXiv:2104.08894},
  year={2021}
}

@article{bac2021scikit,
  title={Scikit-dimension: a python package for intrinsic dimension estimation},
  author={Bac, Jonathan and Mirkes, Evgeny M and Gorban, Alexander N and Tyukin, Ivan and Zinovyev, Andrei},
  journal={Entropy},
  volume={23},
  number={10},
  pages={1368},
  year={2021},
  publisher={MDPI}
}

@article{ma2022delving,
  title={Delving into semantic scale imbalance},
  author={Ma, Yanbiao and Jiao, Licheng and Liu, Fang and Li, Yuxin and Yang, Shuyuan and Liu, Xu},
  journal={arXiv preprint arXiv:2212.14613},
  year={2022}
}

@article{foret2020sharpness,
  title={Sharpness-aware minimization for efficiently improving generalization},
  author={Foret, Pierre and Kleiner, Ariel and Mobahi, Hossein and Neyshabur, Behnam},
  journal={arXiv preprint arXiv:2010.01412},
  year={2020}
}

@article{izmailov2018averaging,
  title={Averaging weights leads to wider optima and better generalization},
  author={Izmailov, Pavel and Podoprikhin, Dmitrii and Garipov, Timur and Vetrov, Dmitry and Wilson, Andrew Gordon},
  journal={arXiv preprint arXiv:1803.05407},
  year={2018}
}

@article{wang2022visual,
  title={Visual recognition with deep nearest centroids},
  author={Wang, Wenguan and Han, Cheng and Zhou, Tianfei and Liu, Dongfang},
  journal={arXiv preprint arXiv:2209.07383},
  year={2022}
}

@inproceedings{lu2023transflow,
  title={Transflow: Transformer as flow learner},
  author={Lu, Yawen and Wang, Qifan and Ma, Siqi and Geng, Tong and Chen, Yingjie Victor and Chen, Huaijin and Liu, Dongfang},
  booktitle={Proceedings of the IEEE/CVF conference on computer vision and pattern recognition},
  pages={18063--18073},
  year={2023}
}

@article{liang2023clustseg,
  title={Clustseg: Clustering for universal segmentation},
  author={Liang, James and Zhou, Tianfei and Liu, Dongfang and Wang, Wenguan},
  journal={arXiv preprint arXiv:2305.02187},
  year={2023}
}

\AppendicesPart

\etocsettocstyle{\section*{Table of Contents}}{}
\etocsetnexttocdepth{subsection}
\localtableofcontents

\appendix

\section{Proof Sketch for the Theoretical Motivation}
\label{sect:proof}

Let the total support of the data $\mathcal{X}$ be the union of $K$ disjoint compact Riemannian sub-manifolds, where each manifold $\mathcal{M}_k$ corresponds to a class indexed by $k$:
\begin{equation}
    \mathcal{X} = \bigcup_{k=1}^{K} \mathcal{M}_k.
\end{equation}
Each class manifold $\mathcal{M}_k$ has its own distinct intrinsic dimension $d_k$, volume $V_k$, and prior probability $\pi_k$.

\subsection{Decomposition of Generalization Error}

We aim to learn a classifier $f: \mathcal{X} \to \{1, \dots, K\}$. The global risk (error) $R(f)$ can be decomposed into the conditional risks for each class manifold:
\begin{equation}
    R(f) = \sum_{k=1}^{K} \pi_k \cdot R_k(f),
\end{equation}
where $R_k(f) = \mathbb{E}_{x \sim \mathcal{M}_k} [\mathbb{I}(f(x) \neq k)]$ is the generalization error specifically on class $k$.

To achieve a good global performance (and specifically, good performance on the minority class), we must bound the risk $R_k(f)$ for every class $k$.

\subsection{Covering Number and Manifold Intrinsic Dimensionality}

Narayanan \& Niyogi \cite{narayanan2009sample} showed that the complexity of learning the decision boundary restricted to the manifold $\mathcal{M}_k$ is governed by the covering number $\mathcal{N}(\epsilon, \mathcal{H}|_{\mathcal{M}_k})$ of that specific manifold. The paper states that the number of balls of radius $\epsilon$ required to cover the space of smooth boundaries on a $d$-dimensional manifold is related to the manifold's volume and dimension:

\begin{lemma}[Component-wise Complexity]
The metric entropy of the hypothesis class restricted to the $k$-th manifold $\mathcal{M}_k$ scales with its specific intrinsic dimension $d_k$:
\begin{equation}
    \log \mathcal{N}(\epsilon, \mathcal{H}|_{\mathcal{M}_k}) \propto V_k \cdot \left(\frac{1}{\epsilon}\right)^{d_k}.
\end{equation}
\end{lemma}

\subsection{Local Metric Entropy and Sample Complexity}

Using standard Uniform Convergence bounds (e.g., VC or Rademacher), the number of samples $n_k$ drawn from class $k$ required to bound the error $R_k(f) \le \epsilon$ with high probability is:
\begin{equation} \label{eq:local_complexity}
    n_k \gtrsim \Omega\left( \left(\frac{1}{\epsilon}\right)^{d_k} \right).
\end{equation}

\section{Extended Literature}
\label{sect:literature}

\subsection{Literature on Long-Tailed Visual Recognition}
\label{sect:literature_long_tail}

Long-tailed recognition approaches can be broadly analyzed under the following main categories \citep{zhang2023deep, yang2022survey, zhang2021bag}: 

\noindent\textbf{Resampling Methods} modify the sampling probabilities of examples  \citep{drummond2003c4}, mainly focusing on under-sampling majority classes or over-sampling minority classes. Notably, class-balanced (uniform) sampling \citep{mahajan2018exploring, kang2019decoupling} gives each class an equal probability of being sampled. To overcome the drawback of overfitting to tail classes in a class-balanced setting, various methods are proposed, such as soft-balancing \citep{peng2020large}, progressive balancing \citep{kang2019decoupling}, and context-based augmentation \citep{Shi2023HowRH}.

\noindent\textbf{Loss Reweighting Methods} adjust the weights of the objective function during training to prioritize less representative classes. Many approaches propose reweighting strategies relying on class frequency \citep{japkowicz2002class, cao2019learning, fernando2022dynamical, zhang2021distribution}. A prominent example by \citet{cui2019class} is the class-balanced loss, which computes loss weights using the effective number of samples for each class. Similarly, \citet{Park2021ibloss} accounts for the influence \citep{cook1982residuals} of samples on decision boundary for reweighting. Another well-known example, Focal loss \citep{lin2017focal}, applies loss reweighting to each example depending on the relative difficulty (based on its prediction probability) of that example during training. There also exist methods that leverage meta-learning \citep{ren2018learning, shu2019meta, jamal2020rethinking} to estimate loss weights from a balanced meta-data.

\noindent\textbf{Margin (Logit) Adjustment Methods} enforce margins in logit space to obtain a balanced decision boundary \citep{ren2020balanced, hong2021disentangling}. Common approaches \citep{cao2019learning, menon2021longtail, Li2022GCL} define these margins over class distributions. LGLA \citep{taoLGLA2023} applies logit adjustment to experts in an ensemble for balanced global and local margins. Differing from these, DRO-LT \citep{samuel2021distributional} enforces a margin in feature space to learn representative features.

\noindent\textbf{Other Approaches} with alternative perspectives are also present: E.g., residual fusion of specialized heads \citep{cui2022reslt}, exploration of an optimal combination of existing techniques \citep{zhang2021bag, Li_2024_CVPR}, a specialized activation function \citep{alexandridis2024adaptive} or distribution-aware expert learning \citep{zhao2023mdcs, wang2020long, cai2021ace, zhang2022self}. Moreover, model calibration techniques to overcome over-confidence show promising improvements \citep{zhong2021improving, zhang2021distribution}. Complementarily, a branch of works focuses on learning unbiased feature representations, with supervised contrastive learning \citep{cui2021parametric, zhu2022balanced, cui2023generalized, li2022targeted} or data augmentation \citep{du2023global}.
There also exist other approaches on visual recognition, such as those based on clustering and class prototypes of feature spaces \cite{wang2022visual, liang2023clustseg, lu2023transflow}.

While this paper presents a brief introduction, there are other families of methods not addressed in this section. For a comprehensive overview, we refer readers to \citep{zhang2023deep, yang2022survey, zhang2021bag}.

\subsection{Literature on Intrinsic Dimension Estimation}
\label{sect:literature_ID}

Intrinsic dimensionality analysis focuses on quantifying the intrinsic qualities of data.
As a general definition, the intrinsic dimension is the minimum number of parameters required to represent a data distribution with minimal loss of generality \citep{fukunaga2013introduction}.
ID is closely related to the manifold hypothesis \citep{gorban2018correction}, where it is assumed that natural image datasets with high dimensionality lie on a manifold of lower dimensionality, which makes learning feasible.
In this context, ID refers to the number of dimensions of this lower-dimensional manifold. The ID of a dataset can be estimated globally (i.e., for the entire dataset), or locally (around a local neighborhood), depending on the scope and the estimation method.
ID estimation methods in the literature can be studied in two general categories: Projective methods \citep{bishop1998bayesian,albergante2019estimating,roweis2000nonlinear,tenenbaum2000global}  and geometric (topological) methods \citep{levina2004maximum, amsaleg2019intrinsic, facco2017estimating}. 

\noindent\textbf{Projective methods}
project data points into lower-dimensional spaces and minimize the projection error in the resulting subspace to estimate the intrinsic dimension of the data. Linear projection with some variant of Principal Component Analysis (PCA) \citep{abdi2010principal, fukunaga1971algorithm, bishop1998bayesian} is a common approach that has been used to estimate ID with varying levels of success. 
FisherS \citep{albergante2019estimating} is another projective method that combines PCA with Fisher separability analysis to determine the intrinsic dimension. Other nonlinear projective methods such as LLE \citep{roweis2000nonlinear} and Isomap \citep{tenenbaum2000global} have also been used for ID estimation as well.

\noindent\textbf{Geometric (topological) methods}
 focus on geometric manipulation of the data to estimate the ID of the underlying manifold. One popular branch of geometric methods is nearest-neighbor methods, where ID is determined from the nearest neighbors of each point in the dataset \citep{levina2004maximum, amsaleg2019intrinsic, facco2017estimating}. Levina and Bickel \citep{levina2004maximum} propose MLE, which estimates the ID by
assigning the sampling probability of point-wise neighbors to a Poisson process and maximizing the associated likelihood, and Facco et al. \citep{facco2017estimating} propose TwoNN, which is another nearest-neighbor method that only requires the two nearest neighbors for each point for ID estimation. Alternatively, fractal-based methods associate the distribution of points in the dataset with a fractal structure, after which tools such as the correlation dimension or the Hausdorff dimension can be used to estimate ID \citep{camastra2002estimating}.

ID estimators can also combine the methods described above, such as in the work of \citet{ceruti2014danco}, however, there is no clear-cut process to determine the optimal ID estimator. The challenge of selecting an ID estimator comes from the fact that natural image datasets trivially do not have ground truth ID values. For more on ID estimation and evaluation of ID estimators, we refer to \citep{campadelli2015intrinsic, camastra2016intrinsic}.

A recent work by \citet{ma2024unveiling} focuses on the intrinsic dimension of perceptual manifolds of balanced datasets in the context of model fairness and introduces ID regularization as an end-to-end trainable loss, and similarly the work by  \citet{ansuini2019intrinsic} also studies the ID of the representation of data in the intermediate layers of a neural network. In comparison, we introduce ID as an imbalance measure and show that it can be used for training as a model-free mitigation method in the context of imbalanced datasets.

\section{Pseudocode of Our Approach}
\label{sect:pseudocode}

We provide a high-level algorithm (see Alg.~\ref{alg:id-alg}) to calculate class-wise IDs for multi-class classification using the FisherS \citep{albergante2019estimating} approach, which describes the process underlined in Section 3. Normalized ID estimates computed here can be used with resampling, reweighting and margin-based methods, as described in Section 5.
\begin{algorithm}[hbt!]
\caption{Pseudocode for our ID-based imbalance measure calculation, using FisherS \citep{albergante2019estimating}. The estimated class-wise measures are utilized as described by Sec. 5.}\label{alg:id-alg}
\begin{algorithmic}[1]
\Require
  \Statex $X \in \mathbb{R}^{n \times d}$: Set of $n$ $d$-dimensional samples.
  \Statex $C$: Conditional number.
  \Statex $\boldsymbol{\alpha}_{s}$: Set of increasing $\alpha$ values with $\alpha \in (0, 1)$.
  \Statex $\mathcal{C}$: Set of all classes.
\Ensure 
    \Statex $\hat{d}_{c}, c=1...|\mathcal{C}|$: Array of normalized ID values.  

\For{class $c \in \mathcal{C}$}
\Statex \hskip1.4em \# Apply preprocessing steps
\State Assign $X_{c} \gets \{x \in X \mid \text{$x$ is from class $c$}\}$
\State Normalize $X_{c} \gets X_{c} - \bar{X_{c}}$
\State Compute $V, U, S = \mathrm{PCA}(X_{c})$, where $U$ are the projections of $X_{c}$ onto principal vectors $V$ with explained variances $S$ in decreasing order
\State Select $k$ principal components, where 
$k = \max\{i: S_{1} / S_{i} < C\}$
\State Whiten columns of $U$: $u_{i} \gets u_{i} / \sigma(u_{i}), i=1...k$
\State Project to unit sphere: $u_{i} \gets u_{i} / ||u_{i}||, i=1...k$
\Statex \hskip1.4em \# Determine optimal $\alpha$ for class $c$
\For{$\alpha \in \boldsymbol{\alpha}_{s}$}
\State Compute empirical inseparability probabilities $p_{\alpha}$ from the projected data $\mathbf{u}_{1:k}$
\State Compute empirical mean $\bar{p_{\alpha}} \gets \frac{1}{N}\sum_{i=1}^{N}p_{\alpha}$
\State Estimate $n_{\alpha}$ from $\bar{p_{\alpha}}$ using Eq. 3
\EndFor
\State Remove $\alpha$ values from $\boldsymbol{\alpha}_{s}$ with invalid $n_\alpha$ estimates
\State Select $\alpha^{*} \in \boldsymbol{\alpha}_{s}$ closest to $0.9 \times \max{\{\boldsymbol{\alpha}_{s}\}}$
\State Assign ID value $d_{c} \gets n_{\alpha^{*}}$
\EndFor
\State Normalize ID values $\hat{d_{c}} \gets d_{c} / \sum_{c'}d_{c'}$
\end{algorithmic}
\end{algorithm}

\section{Implementation, Training and Experiment Details}
\label{sect:implementation}

\subsection{Dataset Details}
\label{sect:dataset_details}

\underline{CIFAR-LT datasets} are long-tailed versions of the standard CIFAR-10 and CIFAR-100  datasets. We adopt the method of \citet{cao2019learning} in our experiments, following prior work.
These datasets follow an exponential long-tailed distribution curve and can be modified by changing the imbalance ratio (IR) of the dataset, with IR defined as:
\begin{equation}
\text{IR ($\rho$)} = \frac{\text{cardinality of the \textit{most} common class}}{\text{cardinality of the \textit{least} common class}}.
\end{equation}
In the literature, IR values of 10, 50, and 100 are common for the long-tailed variants of CIFAR-LT datasets. 

\noindent\underline{Places-LT \citep{liu2019large}} is a long-tailed version of the Places-2 \citep{zhou2017places} scene recognition dataset. It contains 365 scene classes, with class cardinalities ranging from 5 to 4980 samples per class (with a striking IR of 996). The test set is balanced with 100 samples per class.

\noindent\underline{ImageNet-LT \citep{liu2019large}} is a long-tailed modification of ImageNet (ILSVRC-2012) \citep{ILSVRC15}. We use the long-tailed split defined by Liu et al. with 1000 class categories and class cardinalities ranging from 5 to 1280 in the training set. The test set is balanced with 50 samples for each class category.

\subsection{Implementation and Training Details}
\label{sect:imp_and_training_details}

We use ID-based mitigation as a complementary method to existing methods. For this, we use the same pipeline and parameters used in the original methods, only tuning the learning rate accordingly with ID-based reweighting and resampling.
For ID estimation, we use the methods from the scikit-dimension \citep{bac2021scikit} library. Our experiments with model-free approaches use the Bag of Tricks \citep{zhang2021bag} framework. In further experiments on ablations and state-of-the-art comparisons, we additionally integrate ID into the GLMC \citep{du2023global}, SURE \citep{Li_2024_CVPR} and BCL \citep{zhu2022balanced} methods. For our experiments with Places-LT, we finetune a ResNet-152 \citep{he2016deep} model provided by torchvision following the prior work in the literature \citep{liu2019large, cui2022reslt, cui2023generalized}. 

\subsection{Experiment Details}
\label{sect:experiment_details}

For our experiments using the Bag of Tricks \citep{zhang2021bag} framework, we use the training configurations reported in Table~\ref{tab:bag-of-tricks-training-hyperparameters}. For experiments using Focal Loss \citep{lin2017focal}, we additionally sample $\gamma \in \{1.0, 2.0\}$.

The complete Bag of Tricks multi-stage method with CAM-based sampling of \citep{zhang2021bag} uses mix-up sampling \citep{zhang2017mixup} with $\alpha = 1.0$, which we also use in our Bag of Tricks + ID experiments.
Following \citep{zhang2021bag}, our Bag of Tricks + ID setup consists of a first stage of CE training with default sampling and mix-up augmentation, followed by CAM-based example generation using the trained first stage model.
The second stage follows including the CAM-based samples with ID-based resampling and mix-up augmentation, and a short fine-tuning stage with ID-based resampling without mix-up sampling.
Following \citet{zhang2021bag}, we double the epochs for training stages that use mix-up augmentation to account for slow convergence.

For our experiments with Places-LT, we finetune a ResNet-152 \citep{he2016deep} model provided by torchvision following the prior work in the literature \citep{liu2019large, cui2022reslt, cui2023generalized}. We train the ResNet-152 backbone with learning rate sampled from $\ell_{b} \in \{ .01, .02 \}$, and train the classifier (the final fully-connected layer) with a learning rate sampled from $\ell_{c} \in \{.01, .1, .2\}$. 

For our experiments with ImageNet-LT, we train with a ResNet-10 backbone from scratch in accordance with the Bag of Tricks \citep{zhang2021bag} framework, and switch to ResNet-50 \citep{he2016deep} and ResNeXt-50 \citep{xie2017aggregated} based backbones for comparing against the state-of-the-art methods.

Table~\ref{tab:bcl-training-hyperparameters} shows the training configuration for BCL \citep{zhu2022balanced} + ID on ImageNet-LT.
BCL is trained with a base learning rate of 0.1 and a cosine annealing learning rate scheduler. Following \citep{zhu2022balanced}, we use $\lambda = 1.0$ and $\mu = 0.35$ as the logit compensation and balanced contrastive learning loss weights respectively.

The training configurations of SURE \citep{Li_2024_CVPR} on the CIFAR-LT datasets are listed in Table~\ref{tab:sure-rw-training-hyperparameters}. SURE is trained with sharpness-aware minimization \citep{foret2020sharpness} together with stochastic weight averaging \citep{izmailov2018averaging} (SWA) for increased robustness.
SWA is initialized at 120 epochs with a SWA-specific learning rate of 0.05. Mix-up augmentation is used with the mixup parameter sampled from $\mathrm{Beta}(\beta, \beta)$.

Table~\ref{tab:glmc-training-hyperparameters} shows the training configuration for GLMC \citep{du2023global} on ImageNet-LT and CIFAR-LT datasets. On CIFAR-LT datasets, the model is trained with the SGD optimizer with a momentum of 0.9 with a cosine annealing learning rate scheduler. Mix-up augmentation is used with the mixing parameter sampled from $\mathrm{Beta}(\beta, \beta)$.

Regardless of the specific backbone, all experiments on CIFAR-LT were conducted on a single RTX3090 GPU, Places-LT experiments on two RTX3090 GPUs, and ImageNet-LT experiments were conducted with four RTX3090 GPUs with \texttt{torch == 1.12} and \texttt{scikit-dimension == 0.3.4}.

\begin{table*}[hbt!]
    \centering
    \small
    \begin{tabular}{l | c | c | c}
    \toprule
    \multicolumn{1}{c|}{Dataset:} & CIFAR-LT & Places-LT & ImageNet-LT \\
    \midrule
      \rowcolor{gray!30!}
      \multicolumn{4}{c}{Plug-and-play methods + ID parameters} \\
    \midrule
      Learning Rate   & $\ell \in \{.1, .2, .3\}$ & $\ell_{\mathrm{b}} \in \{.01, .02\},$ & $\ell \in \{.1, .2\}$ \\
                      &                             & $\ell_{\mathrm{c}} \in \{.01, .1, .2\}$ & \\
      Epochs & 200 & 30 & 100 \\
      Batch Size & 128 & 128 & $\{256, 512\}$ \\
      Optimizer & SGD & SGD & SGD \\
      Momentum & 0.9 & 0.9 & 0.9 \\
      Weight Decay & $2 \times 10^{-4}$ &$ 5 \times 10^{-4}$  & $1 \times 10^{-4}$ \\
      LR scheduling & Step & Step & Step \\
      LR reduction factor & .01 & .01 & .01 \\
      LR reduction epochs & $\{120, 160\}$ & $\{16, 24\}$ & $\{60, 80\}$ \\
      \midrule
      Data transforms (Train time) & \multicolumn{3}{c}{Random resized cropping, horizontal flipping, color jitter} \\
      Data transforms (Test time) & \multicolumn{3}{c}{Resizing and center cropping} \\
      \midrule
      \rowcolor{gray!30!}
      \multicolumn{4}{c}{Bag of Tricks + ID additional parameters} \\
      \midrule
         LR (1S) & .1 & $\ell_b = .001, \ell_c = .01$ & $\ell \in \{.1, .2\}$ \\
        Epochs (1S) & 320\textsuperscript{\textdagger} & 40\textsuperscript{\textdagger} & 160\textsuperscript{\textdagger} \\
        LR scheduler (1S) & - & multi-step & multi-step \\
        LR scheduler epochs (1S) & - & $\{25, 35\}$ & $\{100, 140\}$ \\
        LR reduction factor (1S) & - & .1 & .1 \\
        \midrule
        LR (2S) & $\ell \in \{.0003, .001\}$ & $\ell_{b} = .001$, & $\ell \in \{.002, .003\}$ \\
                &                                                   & $\ell_{c} \in \{.001, .01\}$ & \\
        Epochs (2S) & 80\textsuperscript{\textdagger} & 20\textsuperscript{\textdagger} & 40\textsuperscript{\textdagger} \\
        LR scheduler (2S) & multi-step & multi-step & multi-step \\
        LR scheduler epochs & $\{40\}$ & $\{8, 16\}$ & $\{30\}$ \\
        LR reduction factor & .01 & .1 & .1 \\
        \midrule
        LR (fine-tuning) & $\ell \in \{.0003, .001\}$ & $\ell_{b} = \{.0003, .001\}$,  & .002 \\
                         &   & $\ell_{c} \in \{.0003, .001\}$ & \\ 
        Epochs (fine-tuning) & 20 & 8 & 20 \\
        LR scheduler epochs & $\{8, 16\}$ & $\{4, 6\}$ & $\{8, 16\}$ \\
        LR reduction factor & .1 & .1 & .1 \\
    \bottomrule
    \end{tabular}
    \caption{Training configurations for experiments using the Bag of Tricks \citep{zhang2021bag} framework. \textsuperscript{\textdagger} indicates training using mixup \citep{zhang2017mixup} augmentation with $\alpha = 1.0$, where we follow \citep{zhang2021bag} in doubling the number of epochs to account for slow convergence. LR: Learning Rate.}
    \label{tab:bag-of-tricks-training-hyperparameters}
\end{table*}

\begin{table}[hbt!]
    \centering
    \small
    \begin{tabular}{l |c}
    \toprule
    \multicolumn{1}{c|}{Dataset:} & ImageNet-LT \\
    \midrule
        Learning Rate & 0.1 \\
        Batch Size & 256 \\
        Epochs & $\{90,  180\}$ \\
        Cosine classifier & \checkmark \\
        Weight Decay & $5 \times 10^{-4}$ \\
        Contrastive Views & RandAug - RandAug \\
        Contrastive Learning Temperature & 0.07 \\
        Optimizer & SGD \\
        Momentum & 0.9 \\
        LR Scheduler & Cosine annealing \\
    \bottomrule
    \end{tabular}
    \caption{Training configuration for BCL \citep{zhu2022balanced} + ID on ImageNet-LT. }
    \label{tab:bcl-training-hyperparameters}
\end{table}
\begin{table}[hbt!]
    \centering
    \small
    \begin{tabular}{l | c} %
    \toprule
    \multicolumn{1}{c|}{Dataset:} & CIFAR-LT \\
    \midrule
    \rowcolor{gray!30!}
    \multicolumn{2}{c}{First stage training} \\
    \midrule
        Learning rate & 0.1 \\
        Batch size & 128  \\
        Epochs & 200 \\
        Weight Decay & $5 \times 10^{-4}$ \\
        Optimizer & SAM (SGD) \\
        Cosine classifier & \checkmark  \\
        $\lambda_{\mathrm{crl}}$ & 0.0 \\
        $\lambda_{\mathrm{mixup}}$ & 1.0 \\
        $\beta$ & 10 \\
    \midrule
    \rowcolor{gray!30!}
    \multicolumn{2}{c}{Second stage fine-tuning} \\
    \midrule
        Learning rate & 0.005  \\
        Batch size & 128  \\
        Epochs & 50  \\
        Re-weighting type & exponential \\
        Cosine classifier & \checkmark  \\
        $\lambda_{\mathrm{crl}}$ & 0.0 \\ 
        $\lambda_{\mathrm{mixup}}$ & 1.0 \\
        $\beta$ & 10 \\
    \bottomrule
    \end{tabular}%
    \caption{Training configurations for SURE + RW \citep{Li_2024_CVPR} + ID on CIFAR-LT.}
    \label{tab:sure-rw-training-hyperparameters}
\end{table}

\begin{table*}[hbt!]
    \centering
    \small
    \begin{tabular}{l |c | c | c}
    \toprule
    \multicolumn{1}{c|}{Dataset:} & ImageNet-LT & CIFAR-10-LT & CIFAR-100-LT \\
            &             & ($\rho = 100 / 50$)  & ($\rho = 100 / 50$)\\
    \midrule
        Learning rate & 0.1 & 0.01 & 0.01 \\
        Batch Size & 120 & 64 & 64 \\
        Epochs & 135 & 200 & 200 \\
        Beta & 0.5 & 0.5 & 0.5 \\
        Momentum & 0.9 & 0.9 & 0.9 \\
        Weight Decay & $2 \times 10^{-4}$ & $5 \times 10^{-3}$ & $5 \times 10^{-3}$ \\
        Contrast Weight & 10 & 1 & 4 / 6 \\
        Label Weighting & 1.0 & 1.2 & 1.2 \\
    \bottomrule
    \end{tabular}%
    \caption{Training configuration for GLMC \citep{du2023global} + ID on CIFAR-LT and ImageNet-LT.}
    \label{tab:glmc-training-hyperparameters}
\end{table*}

\subsection{Computational Overhead of ID Estimation}
Here, we briefly illustrate the computational overhead of training with our method. 
Our method's computational costs only include the calculation of ID: it does not have any additional costs for all mitigation methods (re-sampling, re-weighting, or re-margining).
Table~\ref{tab:id-computational-overhead} includes a wall-clock based timer for estimating ID on various datasets using the FisherS \cite{albergante2019estimating} estimator, including an estimate of the overhead of ID estimation to total training. Here, our reference training schedule is a two-stage re-sampling approach for each respective dataset, and the overhead corresponds to the ratio of total ID estimation time to training a single model.
We reiterate that ID only needs to be computed once per dataset, rather than being re-computed each training session; thus the computational cost of ID estimation is constant with respect to the length of the training stage.
\begin{table}[hbt!]
    \small
    \centering
    \begin{tabular}{l|c|c|c}
    \toprule
    Dataset & Total Time & Time Per Class & Overhead \\
    \midrule
    CIFAR-10-LT ($\rho = 100$) & 56.1 $\pm$ 1.6s & 5.6s & 0.05 \\
    CIFAR-100-LT ($\rho = 100$) & 10.7s $\pm$ 0.3s & 0.1s & 0.01 \\
    ImageNet-LT & 11m2s $\pm$ 7s & 0.6s & 0.01 \\
    Places-LT & 42m59s $\pm$ 33s & 7.1s & 0.03 \\
    \bottomrule
    \end{tabular}
    \caption{Compute table for ID estimation for various datasets.}
    \label{tab:id-computational-overhead}
\end{table}

\section{Additional Results}
\label{sect:additional_results}

\begin{table*}
 \centering
 \small
\begin{tabular}{l | c c c | c c c }
		\toprule
        \multicolumn{1}{c|}{Dataset:}    & \multicolumn{3}{c}{CIFAR-10-LT} & \multicolumn{3}{c}{CIFAR-100-LT} \\
		\midrule \multicolumn{1}{c|}{Imbalance Ratio ($\rho$):}                                    & 10                              & 50                              & 100            & 10             & 50             & 100            \\
		\midrule CE (baseline)\hfill\                                                   & 87.00                           & 76.29                           & 71.58          & 56.04          & 42.89          & 38.90          \\
		LDAM + DRW \citep{cao2019learning}\hfill\                            & 87.76                           & 81.64                           & 77.19          & 57.57          & 46.44          & 41.83          \\
		Log. Adj. \citep{menon2021longtail}\hfill\                       & -                               & -                               & 77.30          & -              & -              & 44.10          \\
		DRO-LT (learned $\epsilon$) \citep{samuel2021distributional}\hfill\  & \underline{91.02}                           & \underline{85.88}                           & \textbf{82.39} & \textbf{64.02} & \textbf{53.95} & \textbf{48.57} \\
		\midrule LDAM + DRW + ID [\textbf{Ours}]\hfill\                     & 87.81                           & 81.12                           & 76.39          & 57.23          & 45.38          & 41.46          \\
		Log. Adj. + ID ($\tau = \tau$*) [\textbf{Ours}]\hfill\       & -                               & -                               & 74.10          & -              & -              & 41.00          \\
		DRO-LT + ID ($\varepsilon =$ N. ID) [\textbf{Ours}]\hfill\  & 90.56                           & 85.39                           & 81.73          & 63.40          & 53.12          & \underline{48.52}          \\
		DRO-LT + ID ($\varepsilon =$ L. + ID) [\textbf{Ours}]\hfill\   & \textbf{91.10}                  & \textbf{85.91}                  & \underline{82.34}          & \underline{63.69}          & \underline{53.63}          & \underline{48.52}          \\
		\bottomrule
	\end{tabular}
     \caption{\textbf{Experiment 1:} Top-1 accuracy results for simple plug-and-play margin-based methods on long-tailed CIFAR datasets. N: normalized. L: Learned epsilon. Log. Adj.: Logit Adjustment.}
	\label{table:cifar-margin-supplementary}
\end{table*}
\subsection{Rarity-based Analysis on One-Stage Model-Free Approaches}
\label{sect:rarity_analysis}

We further analyze the results of ID-based imbalance mitigation with one-stage model-free methods in a rarity-based grouping and present the results in Table \ref{table:cifar100-id-analysis-detailed-supp-mat}. We sort the 100 classes of CIFAR-100 by decreasing example counts and group the 33/34/33 class partitions as head, middle, and tail classes respectively. ID-based reweighting and resampling both outperform alternative methods, and in particular the results show that ID-based mitigation primarily increases the performance of trained in the middle and tail classes of a dataset.
\begin{table*}[hbt!]
 \centering
 \small
    \begin{tabular}{l |  c c c c | c c c c}
		\toprule \multicolumn{1}{c|}{Imbalance Ratio ($\rho$):} & \multicolumn{4}{c}{50} & \multicolumn{4}{c}{100} \\
		\midrule \multicolumn{1}{c|}{Rarity:}      & Head              & Middle              & Tail              & Average            & Head              & Middle              & Tail              & Average            \\
        \midrule
        \rowcolor{gray!30!}
        \multicolumn{9}{c}{Reweighting-based methods} \\
		\midrule CE (baseline)                              & 67.7          & 42.7          & 18.3          & 42.9          & 65.9          & 39.1          & 11.7          & 38.9          \\
		Focal Loss \citep{lin2017focal}            & \textbf{68.4} & 42.5          & 15.1          & 42.0          & 66.9          & 37.4          & 9.4          & 37.9          \\
		CB + Focal                & 65.2          & 44.1          & \textbf{20.0}          & \textbf{43.1}          & \textbf{67.6} & 37.4          & 10.3          & 38.4          \\
		\midrule ID [\textbf{Ours}]    & 65.8          & 43.6          & 19.3          & 42.9          & 66.3          & 38.2          & 12.1          & 38.9          \\
		ID + CB + Focal [\textbf{Ours}]   & 65.3          & \textbf{44.5}          & 19.0          & 43.0          & 66.8          & \textbf{41.6} & \textbf{14.1} & \textbf{40.8} \\
		\midrule
        \rowcolor{gray!30!}
        \multicolumn{9}{c}{Resampling-based methods}  \\
		\midrule CE (baseline)           & \textbf{67.7} & 42.7          & 18.3          & 42.9          & \textbf{65.9} & 39.1          & 11.7          & 38.9          \\
		CB sampling \citep{mahajan2018exploring}     & 61.6          & 41.3          & 18.1          & 40.4          & 57.2          & 32.5          & 10.4          & 33.4          \\
		PB sampling \citep{kang2019decoupling}     & 66.2          & 45.1          & 19.3          & 43.6          & 64.9          & \textbf{40.3} & 12.1          & 39.1          \\
		\midrule                                                
		ID [\textbf{Ours}]      & 52.5          & 39.7          & 18.3          & 36.9          & 50.8          & 33.4          & 10.2          & 31.5          \\
		PB-ID [\textbf{Ours}]     & 65.8          & \textbf{47.0} & \textbf{22.3} & \textbf{45.1} & 64.9          & 40.2          & \textbf{14.1} & \textbf{39.7} \\
		\bottomrule
	\end{tabular}
 	\caption{\textbf{Experiment 1:} Top-1 accuracy results for one-stage plug-and-play resampling and reweighting
	methods for ResNet-32 on long-tailed CIFAR-100.}
	\label{table:cifar100-id-analysis-detailed-supp-mat}
\end{table*}

\subsection{Additional Analysis on FisherS ID Estimation}
\label{sect:additional_fisher_id}

\begin{figure}
    \begin{subfigure}[b]{0.9\linewidth}
     \centering
       \includegraphics[width=\linewidth]{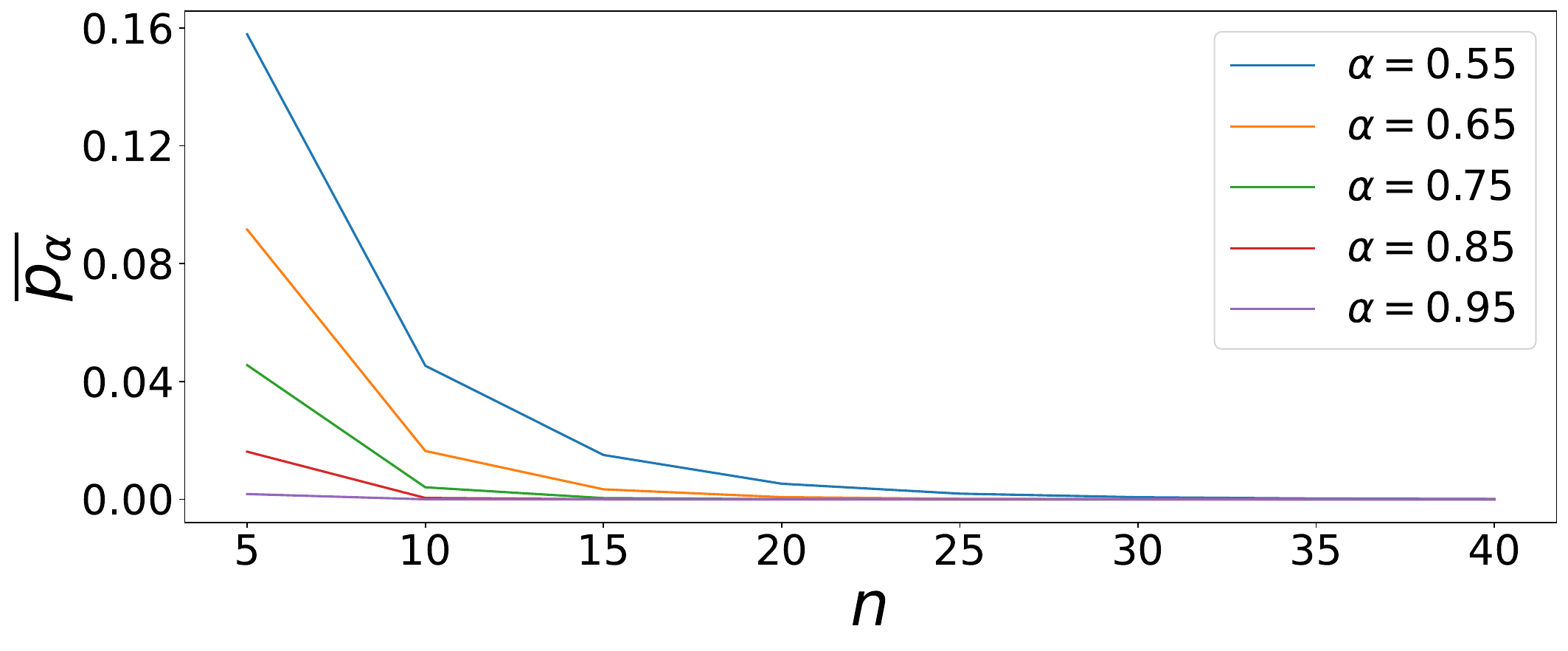}
       \caption{Average inseparability $\overline{p_{\alpha}}$ against increasing dimension $n$.}
       \label{fig:p-alpha-against-n}
    \end{subfigure}
    \begin{subfigure}[b]{0.9\linewidth}
     \centering
       \includegraphics[width=\linewidth]{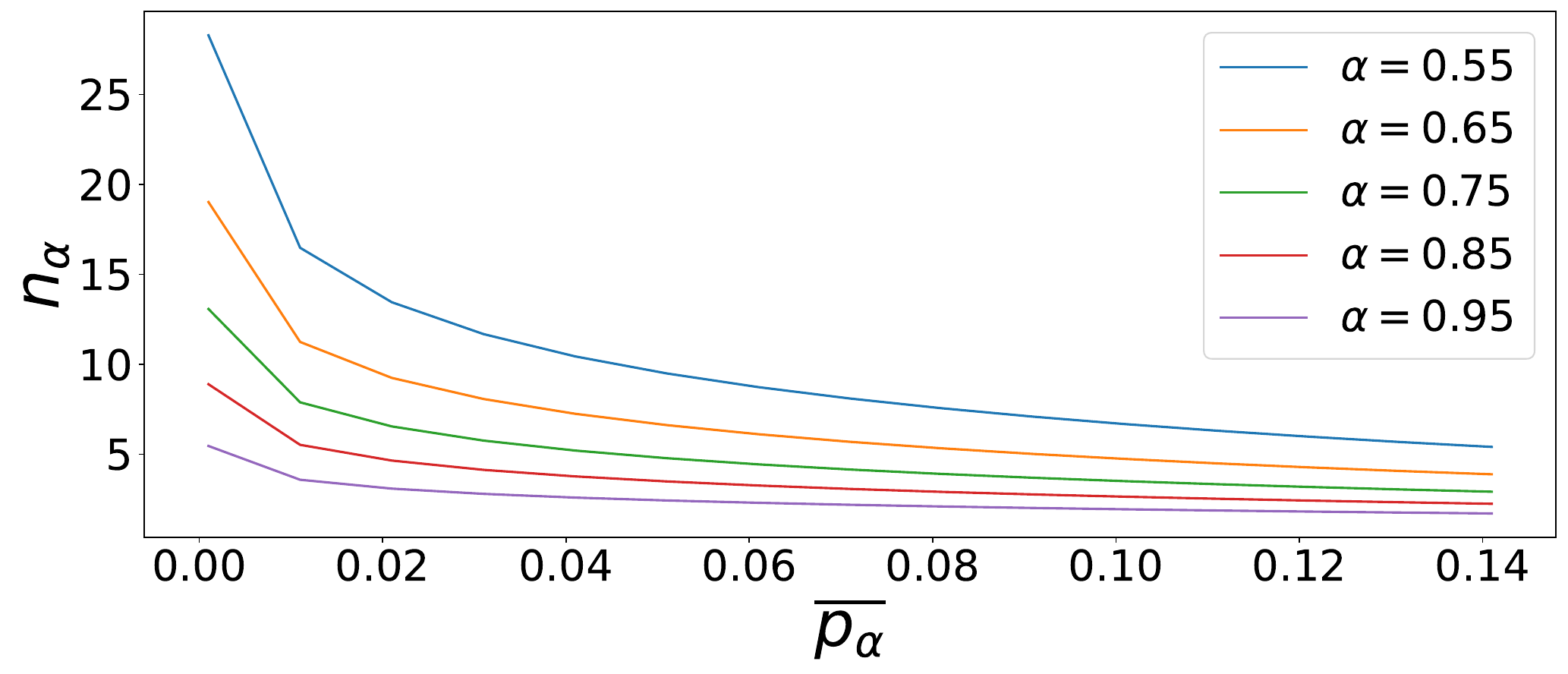}
        \caption{Estimated $n_{\alpha}$ against average inseparability $p_{\alpha}$.}
    \label{fig:n-alpha-against-p-alpha}
    \end{subfigure}
        \caption{Interactions between estimated dimension and degree of separability in FisherS.}
\end{figure}

FisherS \citep{albergante2019estimating} calculates data ID by estimating the mean inseparability of the normalized dataset and determining the candidate dimension for the data which best matches this degree of inseparability for given parameter $\alpha$.
On the equidistribution to a sphere, Fig. \ref{fig:p-alpha-against-n} shows that, as the dimension of the sphere increases, the mean inseparability $\overline{p_{\alpha}}$ on the sphere decreases rapidly following the blessing of dimensionality \citep{gorban2018correction}.
Similarly, given a parameter $\alpha$ and estimated mean inseparability $\overline{p_{\alpha}}$, the estimated intrinsic dimension $n_{\alpha}$ again decreases sharply with increasing inseparability as shown in Fig. \ref{fig:n-alpha-against-p-alpha}.

\begin{figure}
   \centering
   \includegraphics[width=.9\linewidth]{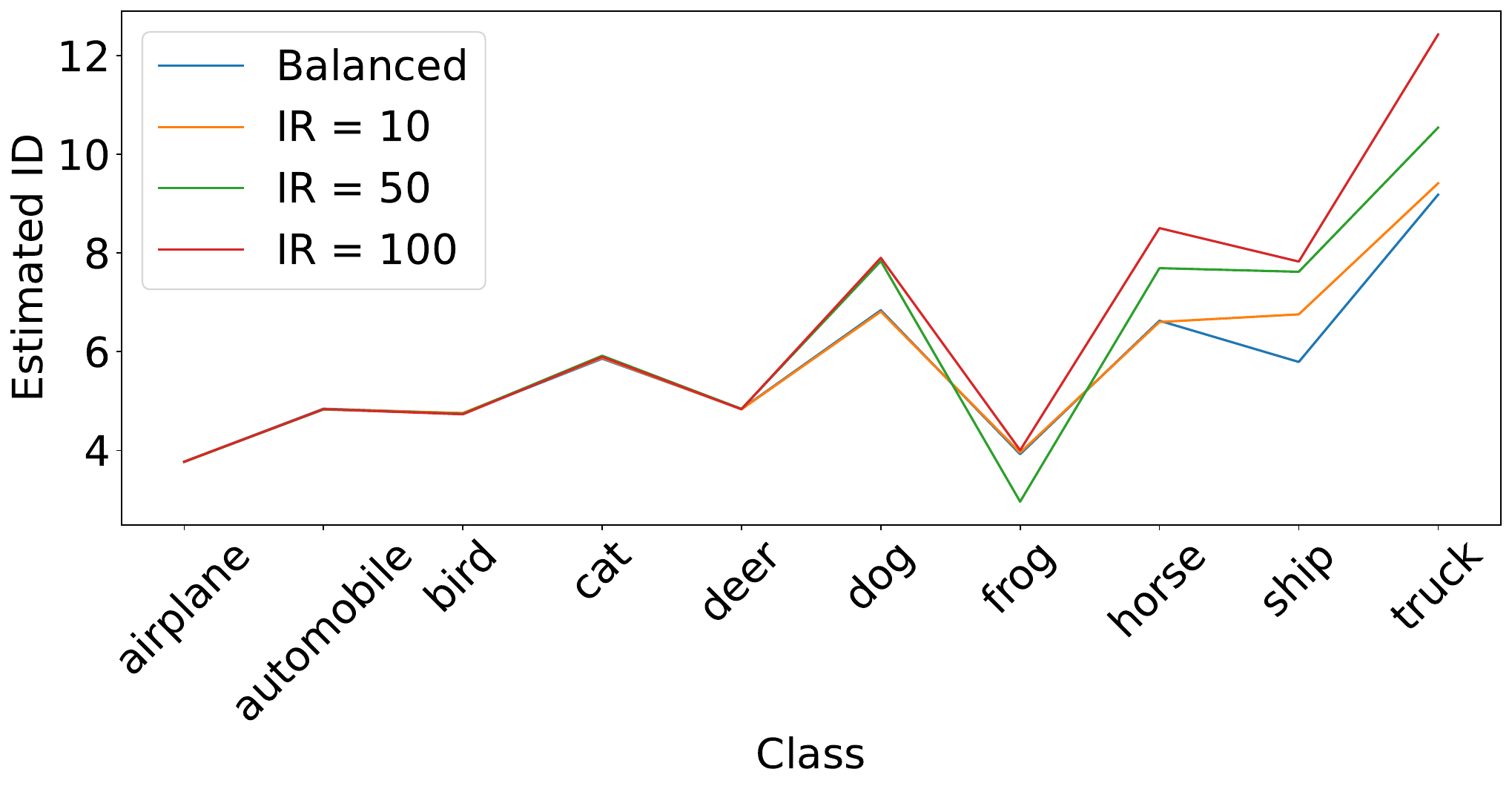}
  \caption{ID estimations of FisherS \citep{albergante2019estimating} under increasing class imbalance for CIFAR-10-LT.}
     \label{fig:fisherS-under-imbalance}

\end{figure}

In addition to the analysis of FisherS as an imbalance measure (in Section~\ref{sec:id}), in Fig. \ref{fig:fisherS-under-imbalance} we plot the ID estimates of classes in the CIFAR-10-LT dataset under increasing cardinality imbalance. We again observe that data ID estimates remain relatively robust under the change in the imbalance ratio.

\begin{figure}[ht!]
    \begin{subfigure}[b]{0.9\linewidth}
     \centering
       \includegraphics[width=\linewidth]{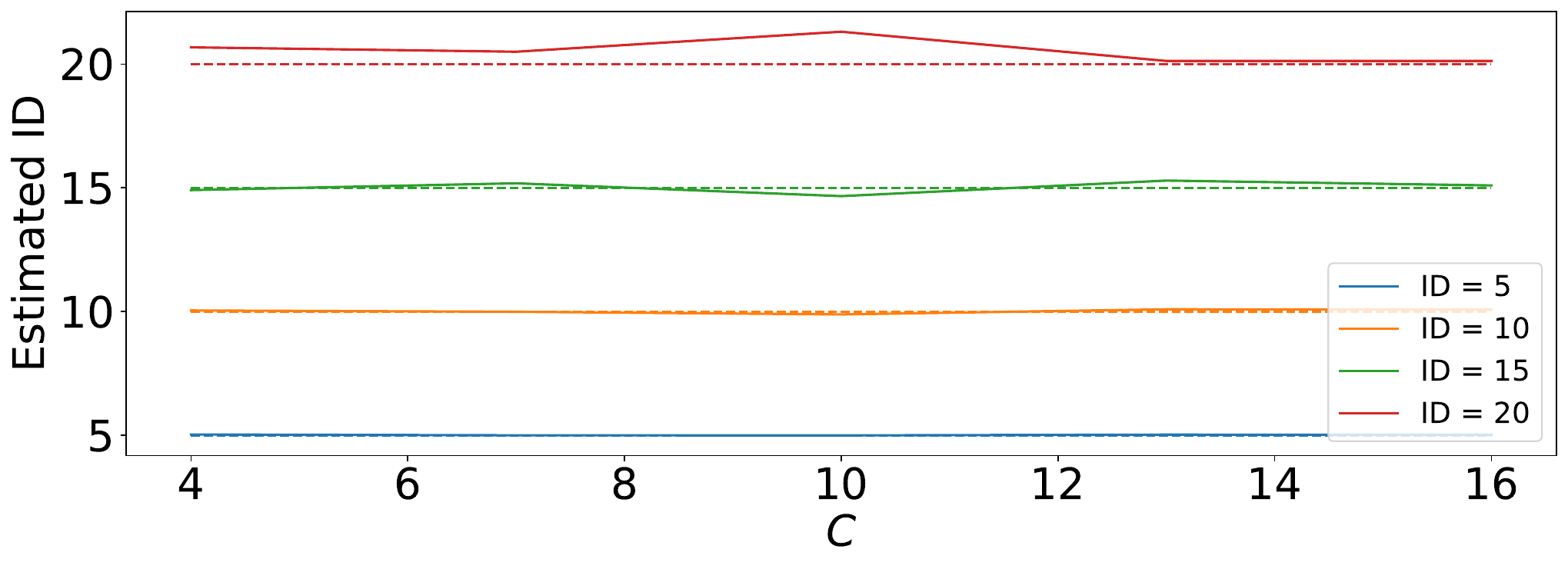}
       \caption{ID estimates with different $C$ values on synthetic multivariate Gaussian data.}
    \end{subfigure}
    \begin{subfigure}[b]{0.9\linewidth}
     \centering
       \includegraphics[width=\linewidth]{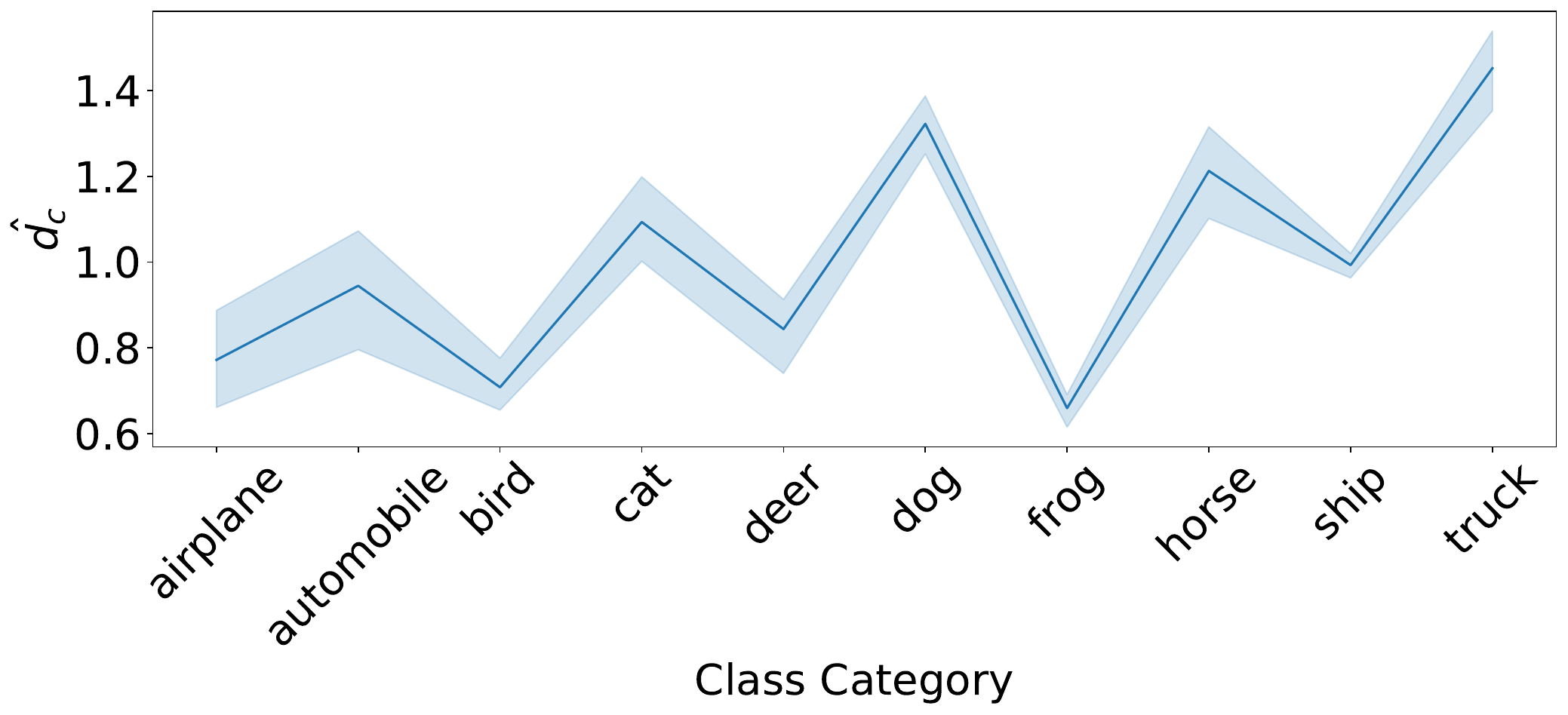}
        \caption{Normalized ID estimates for CIFAR-10 with uniformly sampled $C$ values from $[4, 16]$.}
    \end{subfigure}
        \caption{The sensitivity of ID estimation to $C$.}
        \label{fig:supp-id-against-c}

\end{figure}

Finally, in Fig.~\ref{fig:supp-id-against-c} we analyze the sensitivity of our method equipped with the FisherS estimator to the choice of $C$, controlling the number of components selected in dimensionality reduction. Our findings indicate that our method is robust to changes to $C$ in synthetically generated datasets and the CIFAR-10 dataset, respectively.

\begin{figure}
     \centering
       \includegraphics[width=\linewidth]{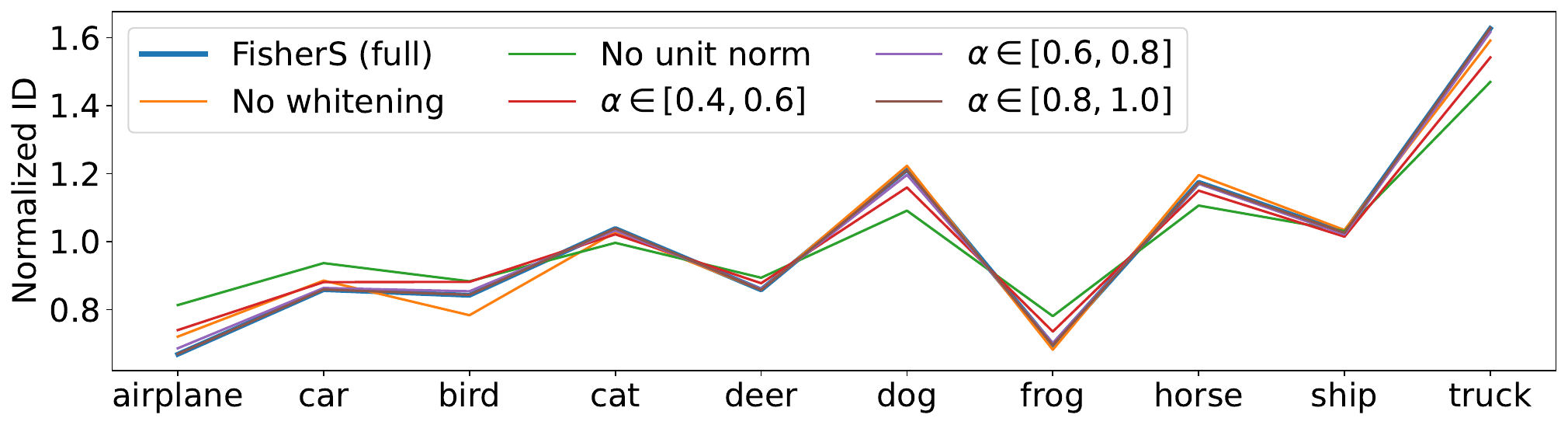}
       \caption{FisherS ID estimates with different hyperparameters on CIFAR-10.}
       \label{fig:id-against-fishers-hyperparameters}
\end{figure}

\subsection{Additional Robustness Analysis of Our Method}
\label{sect:robustness_analysis}

\begin{figure}[hbt!]
    \begin{subfigure}[b]{0.9\linewidth}
     \centering
       \includegraphics[width=\linewidth]{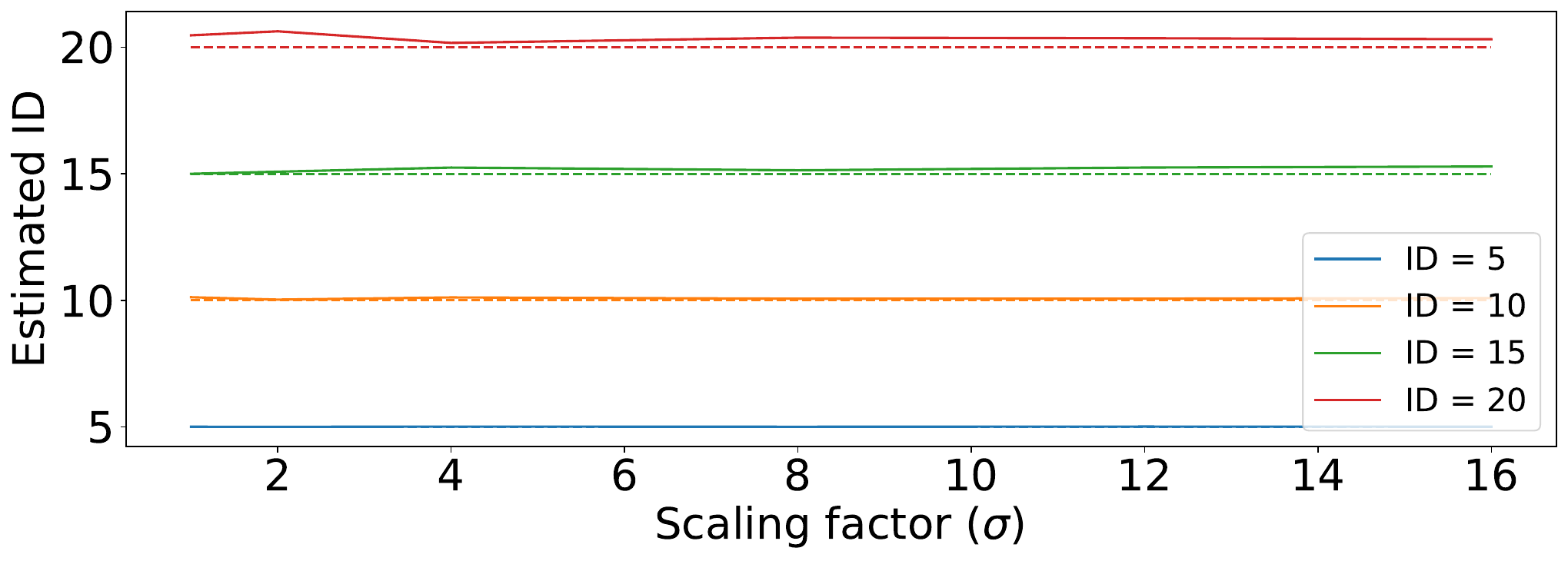}
       \caption{ID estimates with a spherical covariance matrix.}
              \label{fig:id-against-spherical-covariance}

    \end{subfigure}
    \begin{subfigure}[b]{0.9\linewidth}
     \centering
       \includegraphics[width=\linewidth]{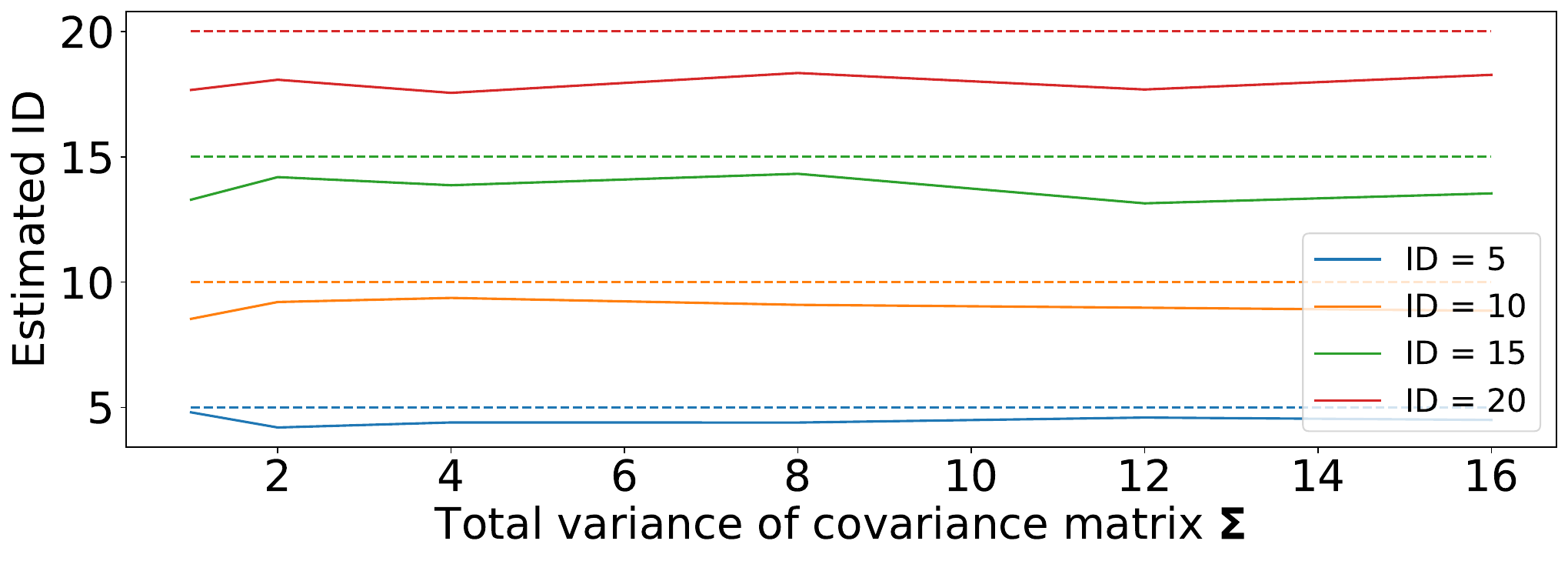}
       \caption{ID estimates with a diagonal covariance matrix.}
    \label{fig:id-against-diagonal-covariance}
    \end{subfigure}
        \begin{subfigure}[b]{0.9\linewidth}
     \centering
       \includegraphics[width=\linewidth]{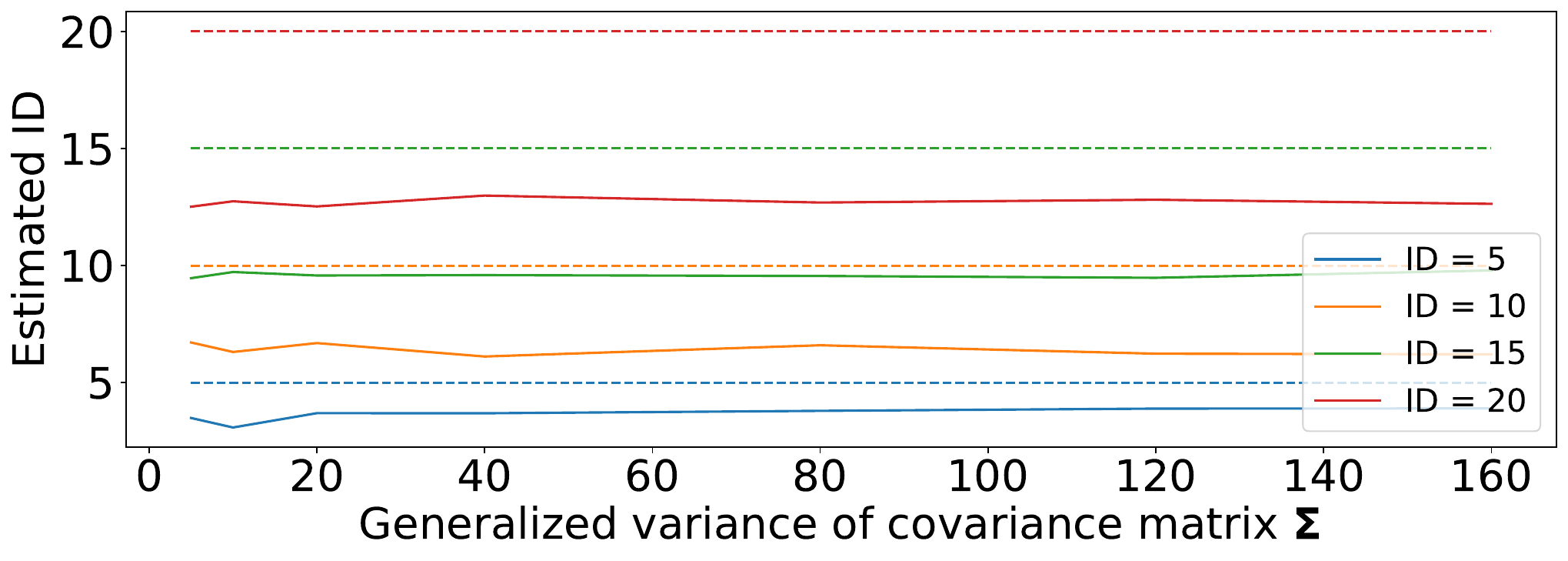}
       \caption{ID estimates with a full covariance matrix.}
    \label{fig:id-against-full-covariance}
    \end{subfigure}
        \caption{ID estimation on synthetic data with known dimensionality with changing variance. 
        Dashed lines indicate the ground truth ID values.}
\end{figure}

\noindent\textbf{Analysis wrt. Dimension Correlation.}
We investigate the effect of increasing variance and correlation between the intrinsic dimensions of the data. For ID = $d$, we sample from multivariate normal distributions $\mathcal{N}(\boldsymbol{\mu}, \mathbf{\Sigma})$ where $\boldsymbol{\mu}$ is a $d$-dimensional zero vector and $\mathbf{\Sigma}$ is a $d \times d$ covariance matrix. In our analysis of the effect of changing variance, we tailor our methods according to the type of covariance matrix:
\begin{itemize}
    \item For normal distributions with spherical covariance matrices ($\mathbf{\Sigma} = \sigma I$), we change the scaling factor $\sigma$ and observe that ID estimation is robust against changing variance, as seen in Fig.~\ref{fig:id-against-spherical-covariance}.
    \item For diagonal covariance matrices ($\mathbf{\Sigma} = \mathrm{Diag}(\boldsymbol{\sigma}_{1, \dots, N})$), we generate diagonal covariance matrices with fixed total variation ($\mathrm{var}_{t} = \mathrm{trace}(\mathbf{\Sigma})$) and observe the change in ID with increasing total variation. As Fig.~\ref{fig:id-against-diagonal-covariance} shows, ID is relatively robust to changing variance in diagonal covariance matrices.
    \item For full covariance matrices, we generate arbitrary full covariance matrices with fixed generalized variance ($\mathrm{var}_{g} = |\mathbf{\Sigma}|$), and observe ID estimates as we gradually increase the generalized variance. In this case, Fig.~\ref{fig:id-against-full-covariance} shows that ID underestimates the actual ID of the data, however, this underestimation is not correlated with increasing generalized variance.
\end{itemize}

\begin{table}[hbt!]
   \small
   \centering
   \begin{tabular}{l | r || c | r}
   \toprule
   CIFAR-10 Class & ID & SVHN Class & ID \\
   \midrule
   airplane & 8.82      & 0 & 4.78 \\
   automobile & 10.83   & 1 & 4.65 \\
   bird & 6.59          & 2 & 4.71 \\
   cat & 10.65          & 3 & 2.98 \\
   deer & 8.75          & 4 & 3.78 \\
   dog & 11.56          & 5 & 2.92 \\
   frog & 9.06          & 6 & 2.86 \\
   horse & 11.74        & 7 & 3.92 \\
   ship & 9.78          & 8 & 5.39 \\
   truck & 13.57        & 9 & 4.67 \\
   \bottomrule
    \end{tabular}
    \caption{\textbf{Experiment 3:} ID estimates for classes in the balanced SVCI-20 dataset with $N = 100$ samples per class.}
    \label{table:id-semantic-imbalance-scores}
\end{table}

\subsection{An Inspection of ID-Cardinality Fusion}
\begin{figure}
    \centering
    \includegraphics[width=0.5\linewidth]{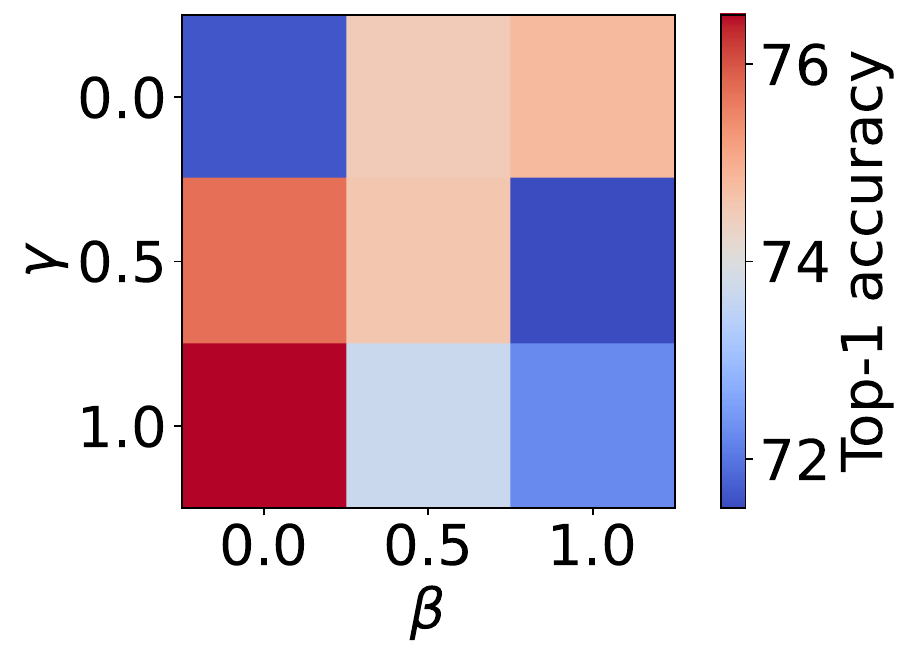}
    \caption{Heatmap of Top-1 accuracies for methods ID-cardinality fusion on CIFAR-10-LT with an imbalance ratio of $\rho = 100$.}
    \label{fig:id-fusion-heatmap}
\end{figure}

In this section, we briefly consider a fusion of ID-based and cardinality-based weighting. Specifically, we introduce a re-weighting strategy with class-based weights $\tilde{w_c}$ defined as the multiplicative combination of FisherS-based ID weighting and class-based weighting: $\tilde{w_c} = N_{c}^{-\beta}d_{c}^{\gamma}$, with scaling parameters $\beta, \gamma \in [0, 1]$.
We then train ResNet-32 models with re-weighting on the CIFAR-10-LT dataset with these fusion of weights.
Fig.~\ref{fig:id-fusion-heatmap} shows the results of these trained models as a heatmap of Top-1 accuracies. In particular, we observe that with these fusion of weights, increasing the importance of the ID-based weighting with parameter $\gamma$ has a larger positive impact on classification accuracy compared to the cardinality-based weighting with parameter $\beta$.

\end{document}